\documentclass{vgtc}                          

\ifpdf
  \pdfoutput=1\relax                   
  \usepackage{graphicx}                
  \DeclareGraphicsExtensions{.pdf,.png,.jpg,.jpeg} 
\else
  \usepackage{graphicx}                
  \DeclareGraphicsExtensions{.eps}     
\fi%

\graphicspath{{figures/}{pictures/}{images/}{./}} 

\usepackage{microtype}                 
\PassOptionsToPackage{warn}{textcomp}  
\usepackage{textcomp}                  
\usepackage{mathptmx}                  
\usepackage{times}                     
\usepackage{cite}                      
\usepackage{tabu}                      
\usepackage{booktabs}                  
\usepackage{amsfonts}
\usepackage{amsmath}
\usepackage{multirow}
\usepackage{tablefootnote}
\usepackage{caption}
\usepackage{subcaption}
\usepackage{transparent}
\usepackage{xcolor}
\usepackage{soul}
\usepackage{hyperref}

\onlineid{8198}

\vgtccategory{Research}

\vgtcinsertpkg

\title{Vox-Fusion: Dense Tracking and Mapping with Voxel-based \\Neural Implicit Representation}

\author{Xingrui Yang$^1$\thanks{Equal contribution} %
\quad Hai Li$^1$\footnotemark[1] %
\quad Hongjia Zhai$^1$ %
\quad Yuhang Ming$^2$ 
\quad Yuqian Liu$^3$
\quad Guofeng Zhang$^1$\thanks{Corresponding author}}
\affiliation{
$^1$State Key Lab of CAD\&CG, Zhejiang University \\
$^2$Visual Information Laboratory, University of Bristol \\
$^3$Autonomous Driving Group, SenseTime \\
{\tt xingruiy@gmail.com, \{garyli, zhj1999, zhangguofeng\}@zju.edu.cn, \\ yuhang.ming@bristol.ac.uk, liuyuqian@senseauto.com}}

\teaser{
  \centering
  \includegraphics[width=\linewidth]{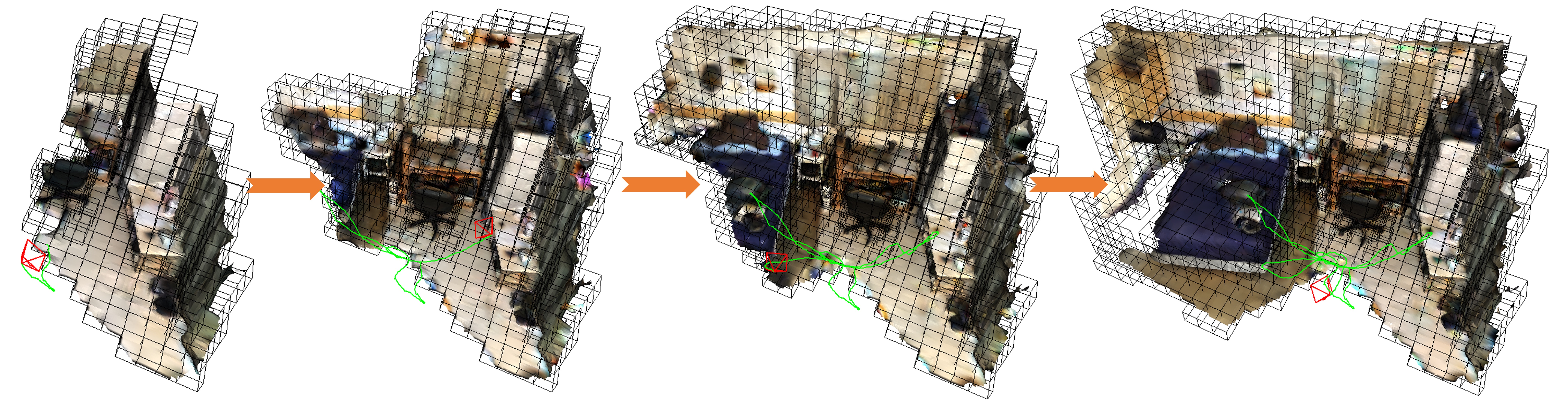}
  \caption{We present Vox-Fusion, a SLAM system that incrementally reconstructs scenes from RGB-D frames using neural implicit networks.  The current camera frame is in red and the camera trajectory is in green. Our method divides the scene into an explicit voxel grid representation (depicted as black bounding boxes) which is gradually built on-the-fly. Inside each voxel, we fuse depth and color observations into local feature embeddings through volume rendering. }
    \label{fig:teaser}
}

\abstract{In this work, we present a dense tracking and mapping system named Vox-Fusion, which seamlessly fuses neural implicit representations with traditional volumetric fusion methods. Our approach is inspired by the recently developed implicit mapping and positioning system and further extends the idea so that it can be freely applied to practical scenarios. Specifically, we leverage a voxel-based neural implicit surface representation to encode and optimize the scene inside each voxel. Furthermore, we adopt an octree-based structure to divide the scene and support dynamic expansion, enabling our system to track and map arbitrary scenes without knowing the environment like in previous works. Moreover, we proposed a high-performance multi-process framework to speed up the method, thus supporting some applications that require real-time performance. The evaluation results show that our methods can achieve better accuracy and completeness than previous methods. We also show that our Vox-Fusion can be used in augmented reality and virtual reality applications. Our source code is publicly available at \url{https://github.com/zju3dv/Vox-Fusion}.
} 

\CCScatlist{
    \CCScatTwelve{Dense SLAM}{Implicit Networks}{Voxelization}{Surface Rendering}
}

\begin{document}

\firstsection{Introduction}
\label{sec:intro}

\maketitle

Dense simultaneous localization and mapping (SLAM) aims to track the 6 degrees of freedom (DoF) poses of a moving RGB-D camera whilst constructing a dense map of the surrounding environment in real-time. It is an essential part of augmented reality (AR) and virtual reality (VR). With high tracking accuracy and the ability to recover complete surfaces, it can support real-time occlusion effects and collision detection during virtual interaction.

Traditional SLAM methods using either feature matching~\cite{newcombe:2010:livedense, orb-slam, orb-slam2, orb-slam3}, nonlinear energy minimization~\cite{newcombe:2011:dtam}, or a combination of both~\cite{dai:2017:bundlefusion, yang:2022:fdslam} to solve the camera poses. These poses are then coupled with their corresponding input point clouds to update a global map represented by geometric primitives such as cost volumes~\cite{newcombe:2011:dtam, weerasekera:2019:gooddense}, surfels~\cite{stuckler:2014:multisurfel, whelan:2015:efusion, wang:2019:densesurfel} or voxels~\cite{newcombe:2011:kinfu, niebner:2013:voxelhashing, kahler:2015:inftam}. Although these methods have been well studied and have shown good reconstruction results, 
they are incapable of rendering novel views as they cannot  hallucinate the unseen parts of the scene. Storing and distributing the maps can be challenging as well due to the requirement of large video memory (VRAM). Moreover, modifying the map on-the-fly is also difficult because of the large number of elements in the map and weaker data associations compared to feature-based methods~\cite{dai:2017:bundlefusion, schops:2019:badslam}.

Focusing on reducing memory usage and improving efficiency, recent works such as CodeSLAM~\cite{bloesch:2018:codeslam} and the follow-up works~\cite{czarnowski:2020:deepfactors, matsuki:2021:codemapping} have demonstrated that neural networks have the ability to encode depth maps using fixed-length optimizable latent embeddings. These latent codes can be updated with multi-view constraints. This method provides a good trade-off between scene quality and memory usage. However, the pre-trained networks used in these systems generalize poorly to different types of scenes, making them less useful in practical scenarios. Also, a consistent global representation is difficult to obtain due to the use of local latent codes.

\begin{figure*}
\begin{center}
\includegraphics[width=0.9\textwidth]{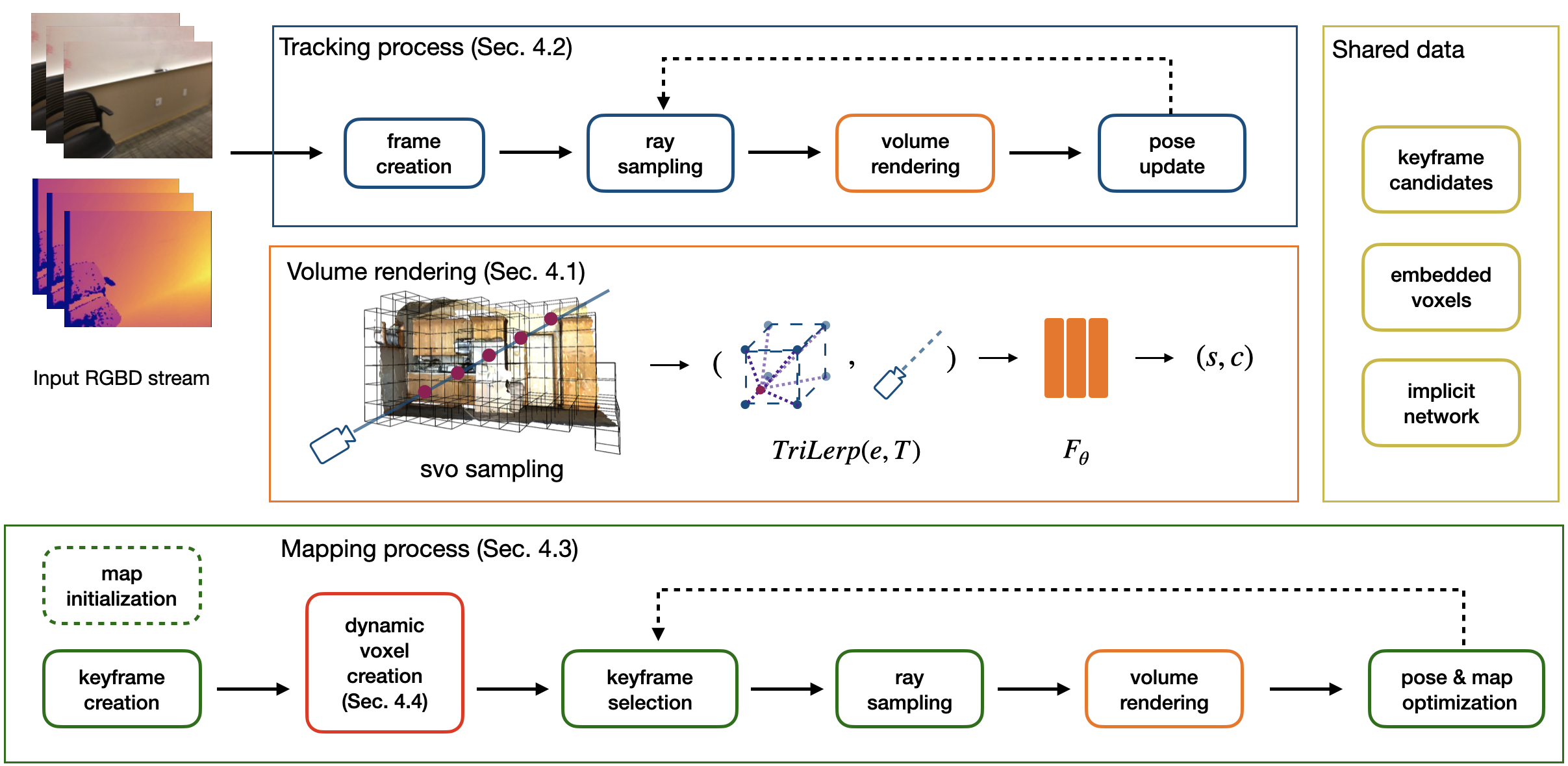}
\end{center}
  \caption{Overview of our SLAM system. Our system consists of three parts: 1) Volume Renderer, which encodes the scene in a MLP and embedding vectors, and outputs the rendered color and SDF value for a given pixel; 2) Tracking Process, which takes as input RGB-D frames and optimizes the camera poses via differentiable rendering and 3) Mapping Process, which reconstructs the geometry of the scene.}
\label{fig:overview}
\end{figure*}

To address these issues, recent works take advantage of the success of NeRF~\cite{mildenhall:2020:nerf} and train a neural implicit network on-the-fly to represent 3D scenes continuously in dense SLAM applications~\cite{sucar:2021:imap, zhu:2021:niceslam}. Specifically, iMap~\cite{sucar:2021:imap} directly uses a single multi-layer perceptron (MLP) to approximate a global scene map and jointly optimizes the map and the camera poses. However, the use of a single MLP makes it difficult to represent geometric details of the scene as well as scale to larger environments without significantly increasing the network capacity.

In this paper, we are interested in mapping unknown scenes with neural implicit networks. Inspired by traditional volumetric SLAM systems~\cite{vespa:2018:supereight} and the successful application of neural implicit representation in parallel tracking and mapping~\cite{sucar:2021:imap}, we propose a more efficient hybrid data structure that combines a sparse voxel representation with neural implicit embeddings. More specifically, we use a sparse octree with Morton coding for fast allocation and retrieval of voxels, which have been proven to be real-time capable for dense mapping~\cite{vespa:2018:supereight}. We model the scene geometry within local voxels as a continuous signed distance function (SDF)~\cite{curless:1996:sdf}, which is encoded by a neural implicit decoder and shared feature embeddings. The shared embedding vectors allow us to use a more lightweight decoder because they contain knowledge of local geometry and appearance. The tracking and mapping process is achieved with differentiable volumetric rendering. We show that our explicit voxel representation is beneficial to AR applications and our mapping method creates more detailed reconstructions compared to current state-of-the-art (SOTA) systems, as we show in the experiments section. To summarize, our contributions are:

\begin{enumerate}
    \item We propose a novel fusion system for real-time implicit tracking and mapping. Our Vox-Fusion combines voxel embeddings indexed by an explicit octree and a neural implicit network to achieve scalable implicit scene reconstruction with sufficient details.
    \item We show that by directly rendering signed distance volumes, our system provides better tracking accuracy and reconstruction quality compared to current SOTA systems with no performance overhead.
    \item We propose to use a fast and efficient keyframe selection strategy based on ratio test and measuring information gain, which is more suitable to maintain large-sized maps.
    \item We perform extensive experiments on synthetic and real-world scenes to demonstrate the proposed method is capable of producing high-quality 3D reconstructions, which can directly benefit many AR applications.
\end{enumerate}

This paper is organized as follows: \autoref{sec:related_work} gives a review of related works. \autoref{sec:overview} presents an overview of our proposed Vox-Fusion system. We explain our reconstruction pipeline in~\autoref{sec:method}. Finally, we evaluate the proposed system on various synthetic and real-world tasks in~\autoref{sec:exp}. We then conclude our paper by introducing potential applications in~\autoref{sec:app} and discussing limitations in~\autoref{sec:limit}.

\section{Related Work}
\label{sec:related_work}
\subsection{Dense SLAM}

\textbf{Traditional SLAM Methods.} DTAM~\cite{newcombe:2011:dtam} introduced the first dense SLAM system that uses every pixel photometric consistency to track a handheld camera. They employ multi-view stereo constraints to update a dense scene model, represented as a cost volume. However, their method is only applicable to small workshop-like spaces. Taking advantage of RGB-D cameras, KinectFusion~\cite{newcombe:2011:kinfu} proposes a novel reconstruction pipeline, which exploits the accurate depth acquisition from commodity depth sensors and the parallel processing power of modern graphics units (GPUs). They track input depth maps with iterative closest point (ICP) and progressively update a voxel grid with aligned depth maps. They also proposed a novel frame-to-model tracking method that greatly reduces short-term drifts with circular camera motion. Following this basic design, many systems gain improvements by introducing different 3D structures~\cite{whelan:2015:efusion}, exploring space subdivision~\cite{roth:2012:mkinfu, niebner:2013:voxelhashing} and performing global map optimization~\cite{kerl:2013:dvo, kahler:2016:inftam2, schops:2019:badslam}. Another interesting research direction is to combine features and dense maps~\cite{dai:2017:bundlefusion, yang:2022:fdslam}, which greatly increases the robustness of iterative methods. These systems provide good results on scene reconstruction at the expense of having a large memory footprint.

\textbf{Learning-based Methods.} Exploiting the power of learned geometric priors, DI-Fusion~\cite{huang:2021:difusion} proposes to encode points in a low dimensional latent space, which can be decoded to generate SDF values. However, the learned geometric prior is inaccurate in complex areas, which leads to poor reconstruction quality. CodeSLAM~\cite{bloesch:2018:codeslam} proposes to use an encoder-decoder structure to embed depth maps as low dimensional codes. These codes, combined with the pre-trained neural decoder, can be used to jointly optimize a collection of key-frames and camera poses. However, similar to other learning-based methods, their approach is not robust to scene variations. Another successful design is to learn feature matching and scene reconstruction separately~\cite{teed:2021:droid}. Recently there have been successful experiments on representing scenes with a single implicit networks~\cite{sucar:2021:imap}. They formulate the dense SLAM problem as a continuous learning paradigm. To bound optimization time, they use heuristic sampling strategies and keyframe selection based on information gain. This method provides a good trade-off between compactness and accuracy. Our method is directly inspired by their design. 

A recent concurrent work~NICE-SLAM~\cite{zhu:2021:niceslam}proposes to tackle the scalability problem by subdividing the world coordinate system into uniform grids. Their system is similar to ours in that we both use voxel features instead of encoding 3D coordinates. However, our approach differs from theirs in the following aspects: (1) NICE-SLAM pre-allocates a dense hierarchical voxel grid for an entire scene, which is not suitable for a practical scenario where the scene bound is unknown, while ours dynamically allocate sparse voxels on the fly, as shown in~\autoref{fig:teaser}, which not only improves usability but also drastically reduces memory consumption; (2) NICE-SLAM uses a pre-trained geometry decoder which could reduce generalization ability while our parameters are all learned on-the-fly, thus our reconstructed surface is not affected by prior information; (3) We propose a keyframe strategy suitable for sparse voxels which is simple and efficient, while NICE-SLAM adopts the strategy from iMap which does not take advantage of the voxel representation.

\subsection{Neural Implicit Networks}

\begin{figure}
    \centering
    \includegraphics[width=.6\linewidth]{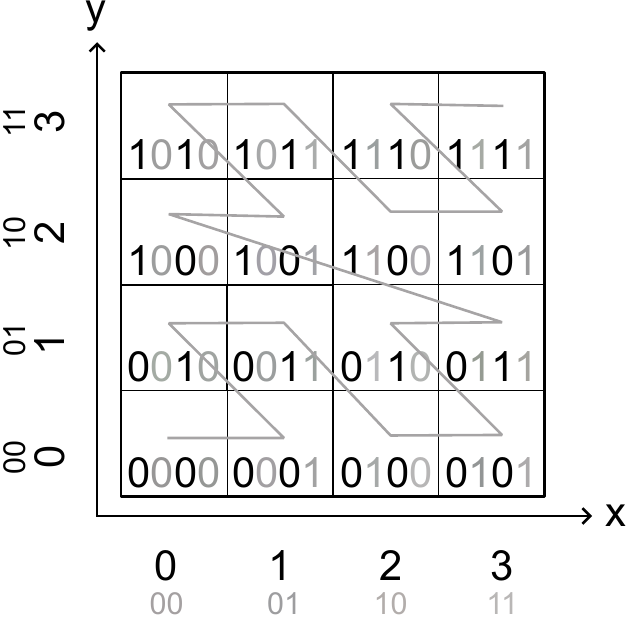}
    \caption{Example of Morton coding on a two-level quad-tree.}
    \label{fig:morton}
\end{figure}

NeRF~\cite{mildenhall:2020:nerf} proposes a method to render the scene as volumes that have density, which is good for representing transparent objects. Their method is only interested in rendering photo-realistic images, a good surface reconstruction is not guaranteed. However, for most AR tasks it is important to know where the surface lies. To solve the surface reconstruction problem, many new methods assume that color is only contributed by points near or on the surface. They propose to identify the surface via iterative root-finding~\cite{oechsle:2021:unisurf}, weighting the rendered color with the associated SDF values~\cite{wang:2021:neus}, forcing the network to learn more details near the surface within a pre-defined truncation distance~\cite{azinovic:2022:neuralrgbd}. We adopt the rendering method from~\cite{azinovic:2022:neuralrgbd} but instead of regressing absolute coordinates, we work on interpolated voxel embeddings.

As a single network often has limited capacity and cannot be scaled to larger scenes without dramatically increasing the number of learnable parameters, NSVF~\cite{liu:2020:nsvf} proposes to embed local information in a separate voxel grid of features. They are able to generate similar if not better results with fewer parameters. Plenoxels~\cite{yu:2022:plenoxels} introduced spherical harmonic functions as voxel embeddings, which completely removed the need for a neural network. The other benefit of an explicit feature grid is the faster rendering speed, since the implicit network can be much smaller compared to the original NeRF network. This uniform grid design can be found in other neural reconstruction methods. E.g. NGLOD~\cite{takikawa:2021:nglod} leverages a hierarchical data structure by concatenating features from each level to achieve scene representation with different levels of details. 

Our hybrid scene representation is inspired by recent works that use voxel-based neural implicit representations~\mbox{\cite{liu:2020:nsvf, Vox-Surf}}. More specifically, we share a similar structure with Vox-Surf~\mbox{\cite{Vox-Surf}} which encodes 3D scenes with neural networks and local embeddings. However, Vox-Surf is an offline system that requires posed images, while our system is a SLAM method that consumes consecutive RGB-D frames for pose estimation and scene reconstruction simultaneously. Vox-surf also needs to allocate all voxels in advance, while our system can manage voxels on-the-fly to allow dynamic growth of the map. Moreover, we use a different rendering function that reconstructs scenes more efficiently.

\section{System overview}
\label{sec:overview}
An overview of our system is shown in~\autoref{fig:overview}. 
The input to our system is continuous RGB-D frames which consist of RGB images $I_i\in \mathbb{R}^3$ and depth maps $D_i \in \mathbb{R}$.
We use the pinhole model as the default camera model and assume that the intrinsic matrix  $K\in \mathbb{R}^{3\times3}$ of the camera is known. 
Similar to other well-known SLAM architectures, our systems also maintain two separate processes: a tracking process to estimate the current camera pose as the frontend and a mapping process to optimize the global map as the backend. 

\begin{figure}
\begin{center}
\includegraphics[width=0.8\linewidth]{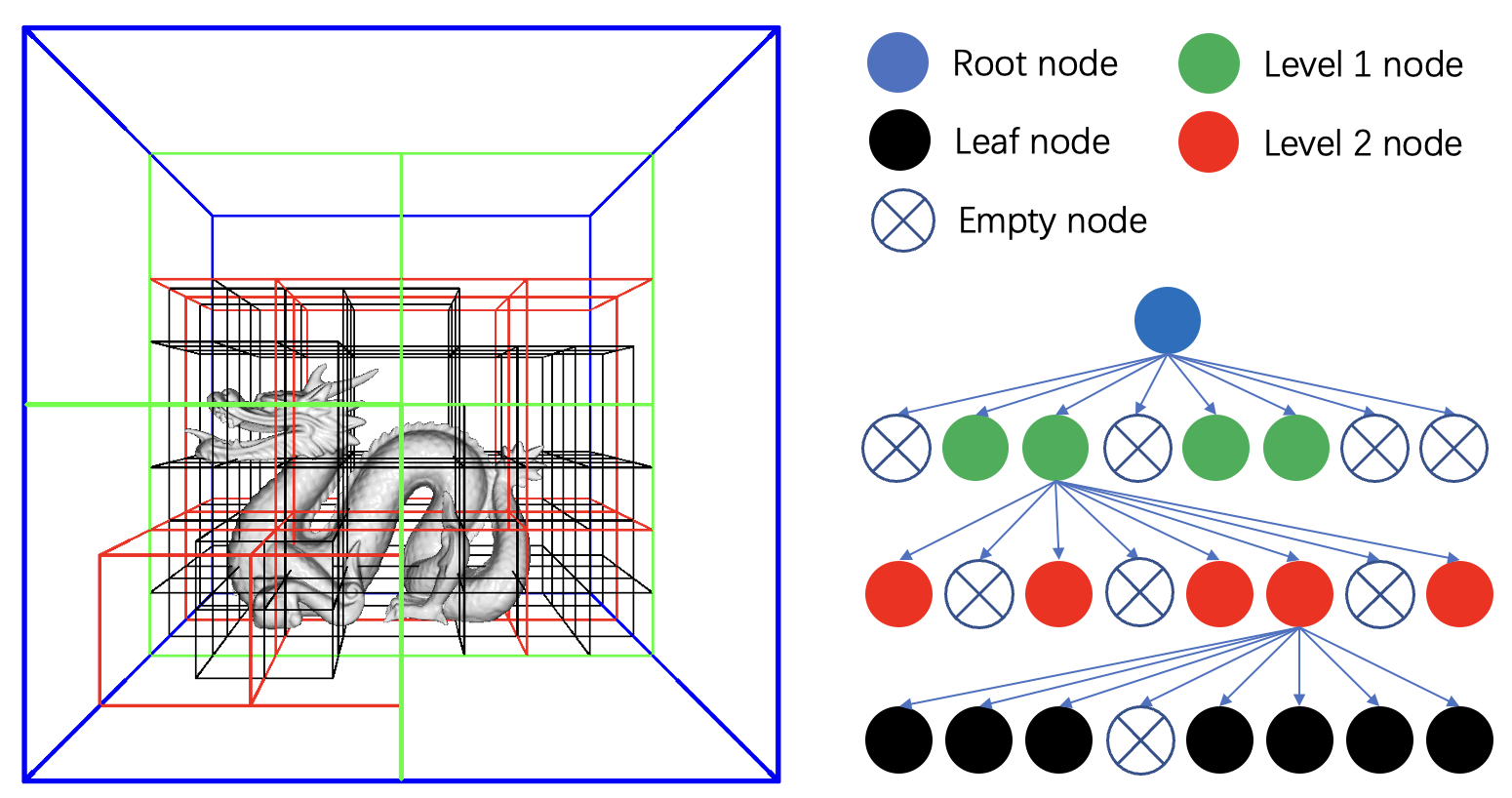}
\end{center}
  \caption{Example of our hierarchical octree structure. For simplicity we only show a four-level octree here.}
\label{fig:octree}
\end{figure} 

When the system starts, we initialize the global map by running a few mapping iterations for the first frame. For subsequent frames, the tracking process first estimate a 6-DoF pose $T_i\in SE(3)$ w.r.t. the fixed implicit scene network $F_{\theta}$ via our differentiable volume rendering method. Then, each tracked frame is sent to the mapping process for constructing the global map. The mapping process first takes the estimated camera poses $T_i$ from frontend and allocates new voxels from the back-projected and appropriately transformed 3D point cloud $P_i\in \mathbb{R}^3$ from the input depth map $D_i$. Then it fuses the new voxel-based scene into the global map and applies the joint optimization. In order to reduce the complexity of optimization, we only maintain a small number of keyframes, which are selected by measuring the ratio of observed voxels. And the long-term map consistency is maintained by constantly optimizing a fixed window of keyframes. 
These individual components will be explained in detail in the following subsections. 

\section{Method}
\label{sec:method}

\subsection{Volume Renderer}
\label{sec:render}
\textbf{Voxel-based sampling.} We represent our scene as an implicit SDF decoder $F_{\theta}$ with optimizable parameters $\theta$, and a collection of $N$-dimensional sparse voxel embeddings. 
The voxel embeddings are attached to the vertices of each voxel and are shared by neighboring voxels. 
The shared embeddings alleviate voxel border artifacts as commonly seen in non-sharing structures~\cite{huang:2021:difusion}. 

However, sampling points inside the sparse voxel representation is not straightforward. Naive sampling methods such as stratified random sampling~\cite{mildenhall:2020:nerf} waste computational power on sampling spaces that are not covered by valid voxels. Therefore, we adopt the method used in~\cite{liu:2020:nsvf} to perform efficient point sampling. For each sampled pixel, we first check if it has hit any voxel along the visual ray by performing a fast ray-voxel intersection test. Pixels without any hit are masked out since they do not contribute to rendering. As the scene could be unbounded in complex scenes, we enforce a limit $M_h$ on how many voxels a single pixel is able to see. Unlike prior works where the limit is heuristically specified~\cite{liu:2020:nsvf, Vox-Surf}, we dynamically change it according to the specified maximum sampling distance $D_{max}$.

\textbf{Implicit surface rendering.} Unlike NeRF~\cite{mildenhall:2020:nerf} where an MLP is used to predict occupancy for 3D points, we directly regress SDF values which is a more useful geometric representation that can support tasks such as ray tracing. The key to our approach is to use voxel embeddings instead of 3D coordinates as opposed to previous works. To render color and depth from the map, we adopt the volume rendering method proposed in~\cite{azinovic:2022:neuralrgbd}. However, we modify it to apply to feature embeddings instead of global coordinates.
We sample $N$ points to render the color for each ray.
More specifically, we use the following rendering function to obtain the color $\textbf{C}$ and depth $D$ for each ray:
\begin{gather}\label{eq:render}
    (\mathbf{c}_i,s_i)=F_{\theta}(\text{TriLerp}(\hat{\xi}_iT_i\mathbf{p}_i, \mathbf{e})), \\
    w_i = \sigma(\frac{s_i}{tr})\cdot \sigma(-\frac{s_i}{tr}), \\ 
    \mathbf{C} = \frac{1}{\sum^{N-1}_{j=0}w_j}\sum^{N-1}_{i=0}w_j\cdot\mathbf{c}_j, \\
    D = \frac{1}{\sum^{N-1}_{j=0}w_i}\sum^{N-1}_{j=0}w_j\cdot d_j, 
\end{gather}
where $T_i$ is the current camera pose, $\text{TriLerp}(\cdot, \cdot)$ is the trilinear interpolation function, $F_{\theta}$ is the implicit network with trainable parameters $\theta$, and $\xi\in \mathfrak{se}(3)$ is the frame pose update. $\mathbf{c}_j$ is the predicted color for each 3D point from the network, by trilinearly interpolating voxel embeddings $\mathbf{e}$. Likewise, $s_i$ is the predicted SDF value, and $d_j$ the $j$-th depth sample along the ray. $\sigma(\cdot)$ is the sigmoid function and $tr$ is a pre-defined truncation distance. The depth map is similarly rendered from the map by weighting sampled distance instead of colors. 

\textbf{Optimization.}
To supervise the network, we apply four different loss functions: RGB loss, depth loss, free-space loss and SDF loss on the sampled points $P$.
The RGB and depth losses are simply absolute differences between render and ground-truth images:
\begin{equation}
\begin{split}
    \mathcal{L}_{rgb} = \frac{1}{\vert P\vert}\sum_{i=0}^{\vert P\vert}\lVert \textbf{C}_i - \textbf{C}^{gt}_i\rVert ,\\ \mathcal{L}_{depth} = \frac{1}{|P|}\sum_{i=0}^{\vert P\vert}\lVert D_i - D^{gt}_i\rVert ,
\end{split}
\end{equation}
where $D_i,C_i$ are the rendered depth and color of the $i$-th pixel in a batch, respectively. $D_i^{gt},C_i^{gt}$ are the corresponding ground truth values. The free-space loss works with a truncation distance $tr$ within which the surface is defined. The MLP is forced to learn a truncation value $tr$ for any points lie within the camera center and the positive truncation region of the surface:
\begin{equation}
    \mathcal{L}_{fs} = \frac{1}{\vert P\vert }\sum_{p\in P} \frac{1}{S_p^{fs}} \sum_{s \in S_p^{fs}}(D_s - tr)^2.
\end{equation}
Finally, we apply SDF loss to force the MLP to learn accurate surface representations within the surface truncation area:
\begin{equation}
    \mathcal{L}_{sdf} = \frac{1}{\vert P\vert}\sum_{p\in P} \frac{1}{S_p^{tr}} \sum_{s \in S_p^{tr}}(D_s - {D}_s^{gt})^2.
\end{equation}
Unlike methods such as~\cite{rematas:2022:urf} that force the network to learn a negative truncation value $-tr$ for points behind the truncation region, we simply mask out these points during rendering to avoid solving surface intersection ambiguities~\cite{wang:2021:neus} as proposed in~\cite{azinovic:2022:neuralrgbd}. This simple formulation allows us to obtain accurate surface reconstructions with a much faster processing speed.

\subsection{Tracking}
\label{sec:tracking}

During tracking, we keep our voxel embeddings and the parameters of the implicit network fixed, i.e., we only optimize a 6-DoF pose $T\in SE(3)$ for the current camera frame. Similar to previous methods where pose estimates are iteratively updated by solving an incremental update, in each update step, we measure the pose update in the tangent space of $SE(3)$, represented as the lie algebra  $\xi \in \mathfrak{se}(3)$. We assume a zero motion model where the new frame is sufficiently close to the last tracked frame, therefore we initialize the pose of the new frame to be identical to that of the last tracked frame. Although other motion models such as constant motion are also applicable here. For each frame, we sample a sparse set of $N_t$ pixels from the input images for tracking. We follow the procedure described in~\autoref{sec:render} to sample candidate points and perform volume rendering, respectively. The frame pose is updated in each iteration via back-propagation. Similar to~\cite{sucar:2021:imap}, we keep a copy of our SDF decoder and voxel embeddings for the tracking process. This map copy is directly obtained from the mapping process and updated each time a new frame has been fused into the map.
\begin{figure}
    \centering
    \includegraphics[width=\linewidth]{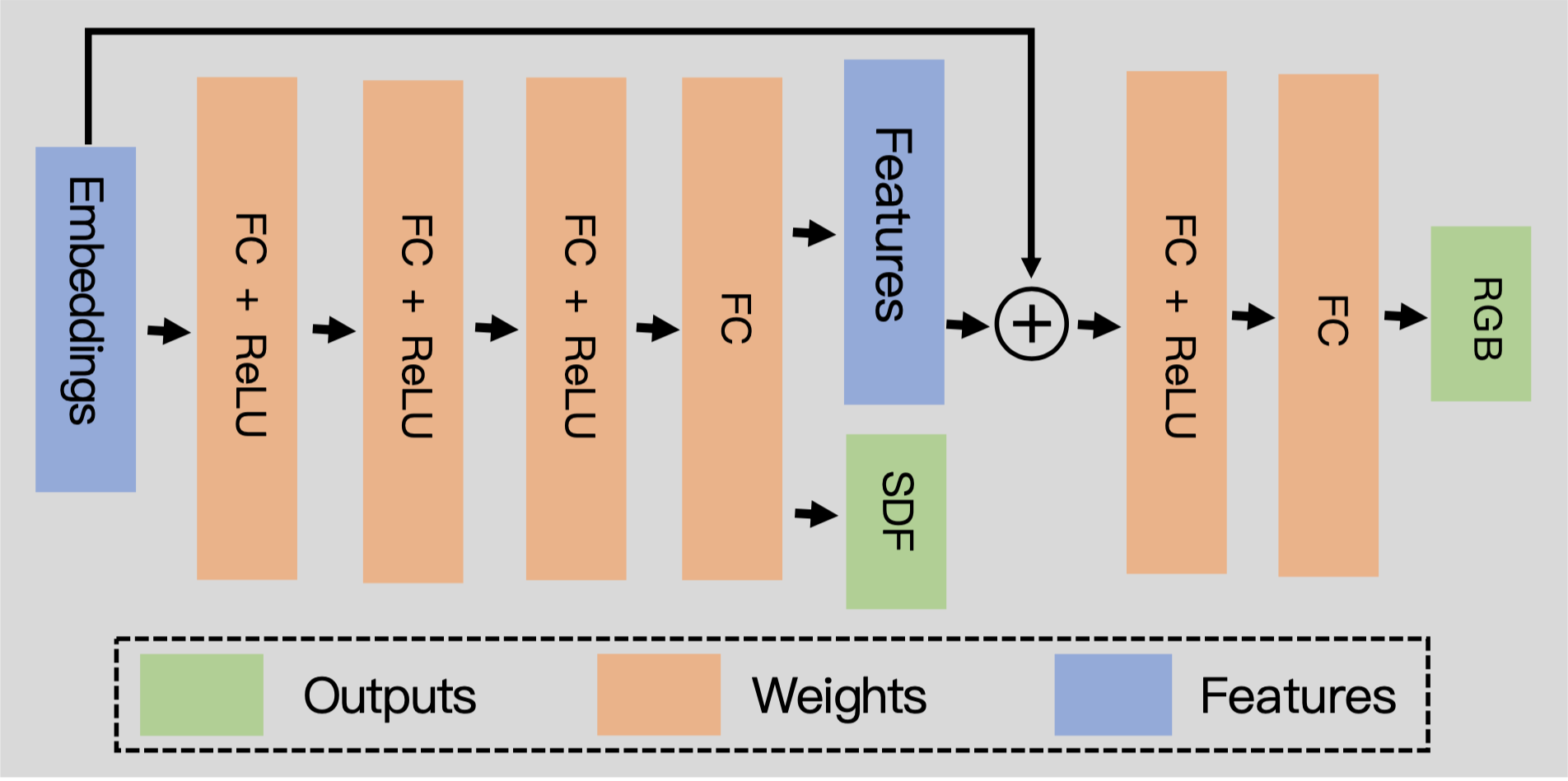}
    \caption{Network architecture.}
    \label{fig:arch}
\vspace{-1em}
\end{figure}

\subsection{Mapping}
\begin{figure*}
    \centering
    \begin{subfigure}[t]{0.33\linewidth}
        \centering
        \includegraphics[width=\textwidth]{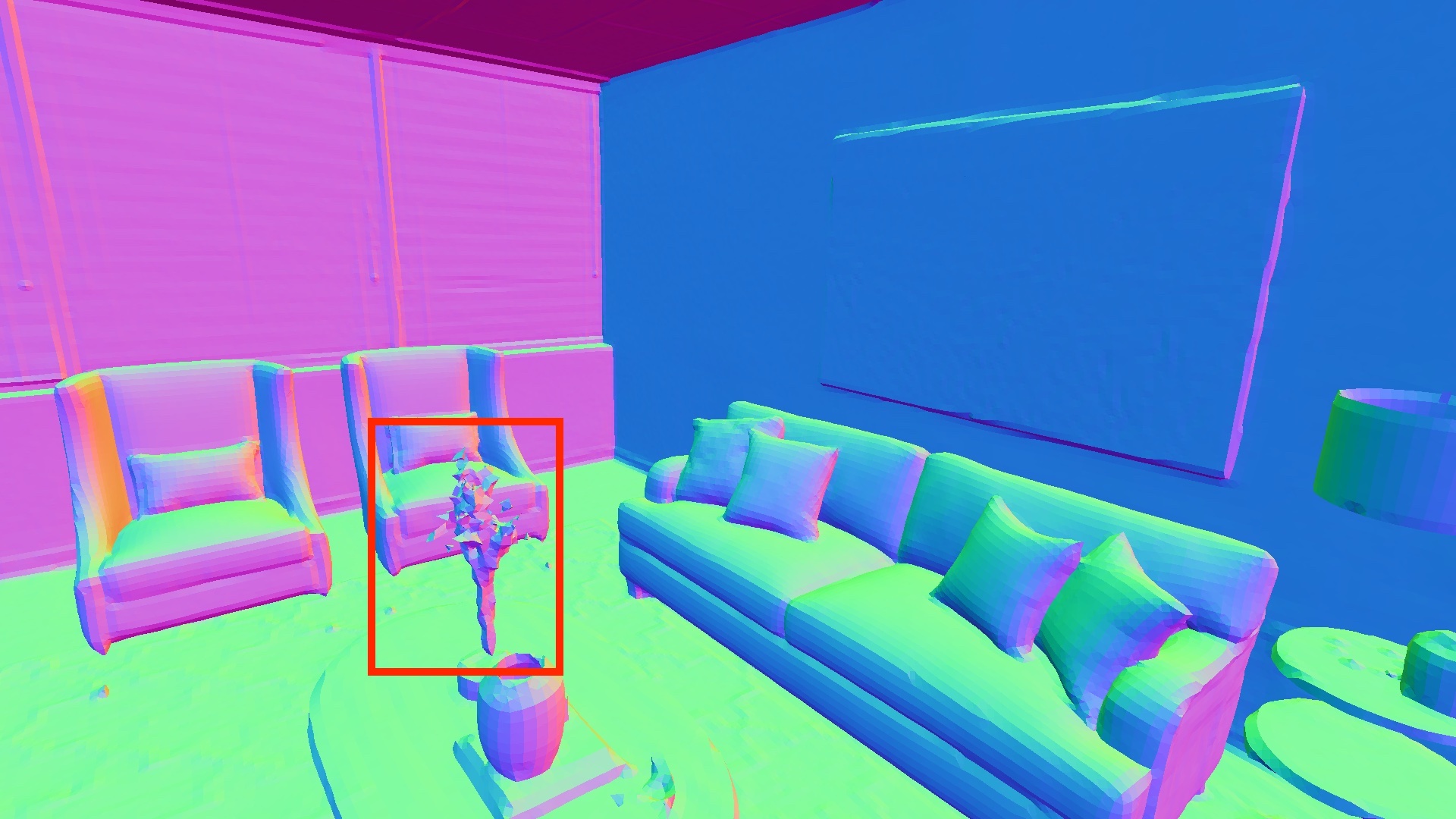}
    \end{subfigure}
    \begin{subfigure}[t]{0.33\linewidth}
        \centering
        \includegraphics[width=\textwidth]{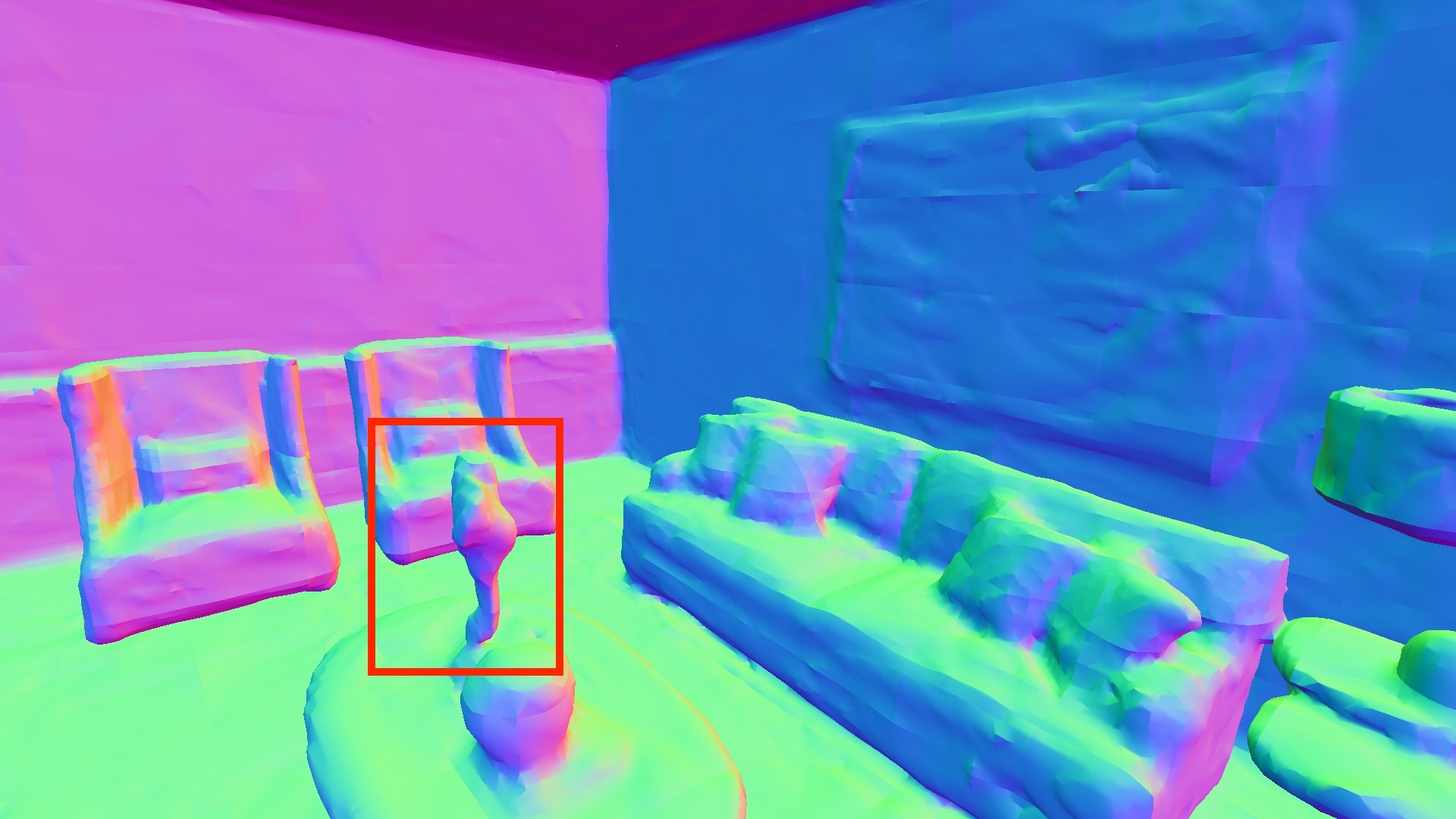}
    \end{subfigure}
    \begin{subfigure}[t]{0.33\linewidth}
        \centering
        \includegraphics[width=\linewidth]{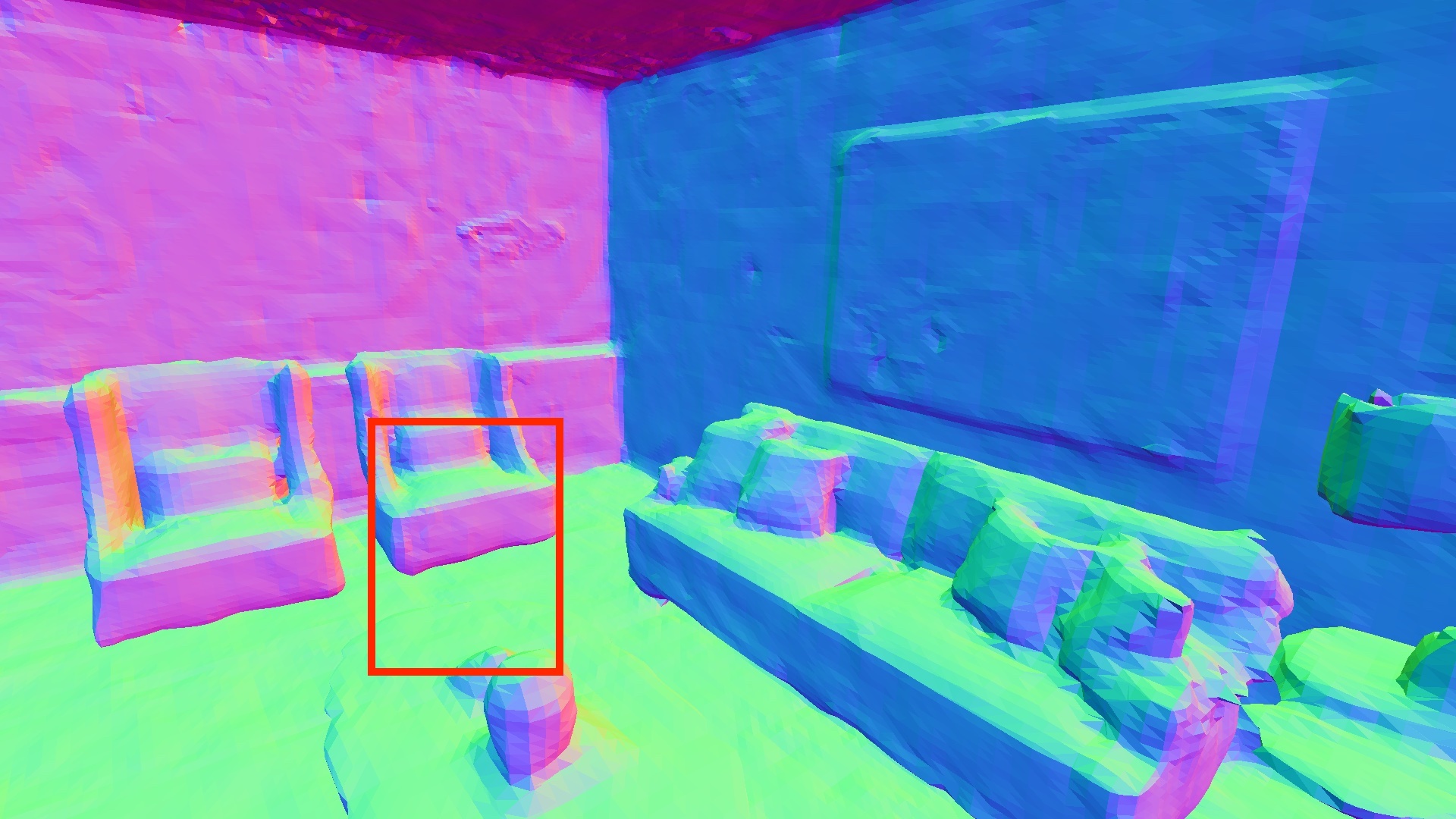}
    \end{subfigure}
    \hfill
    \begin{subfigure}[t]{0.33\linewidth}
        \centering
        \includegraphics[width=\linewidth]{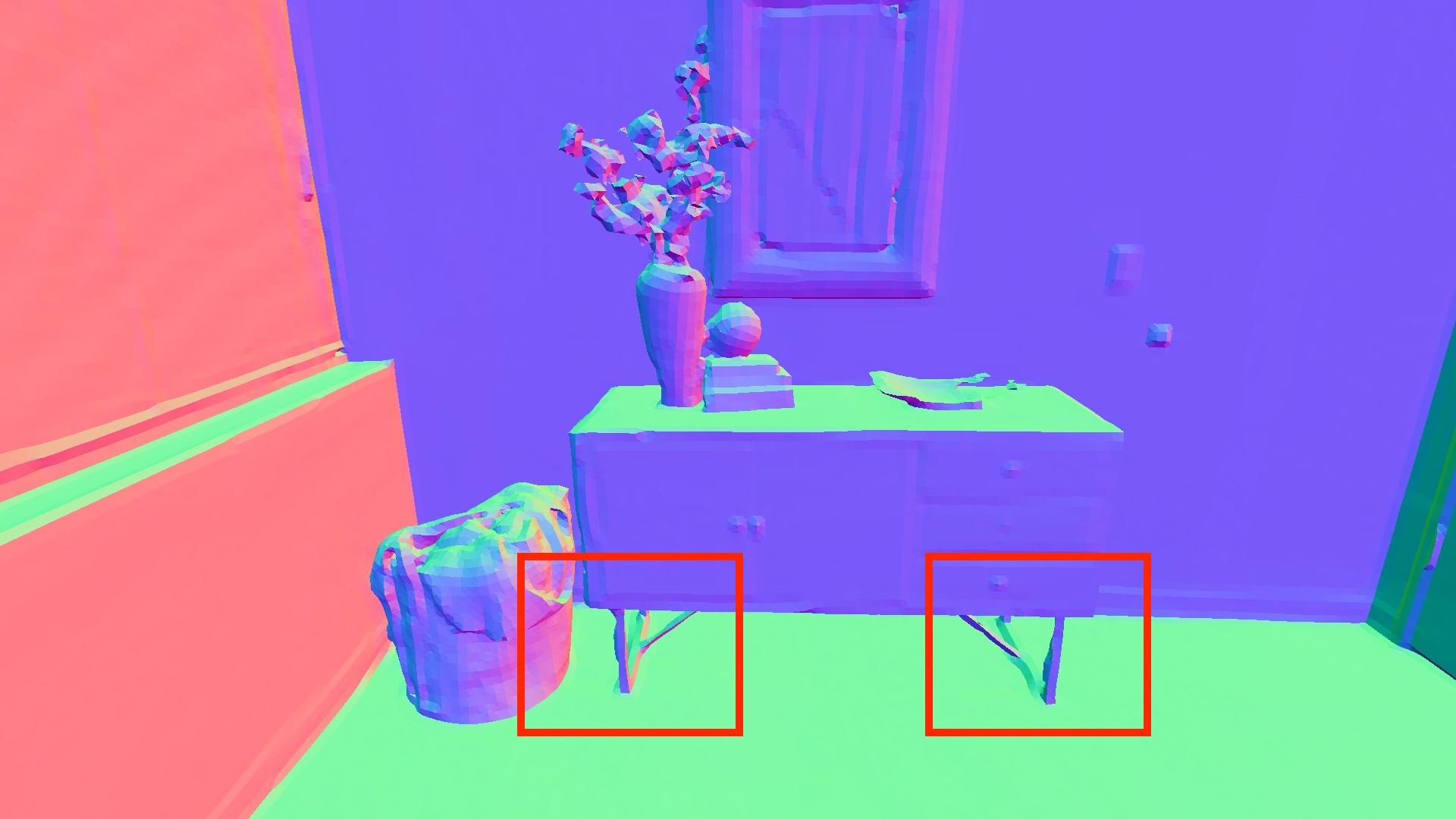}
        \caption{Ground truth}
    \end{subfigure}
    \begin{subfigure}[t]{0.33\linewidth}
        \centering
        \includegraphics[width=\textwidth]{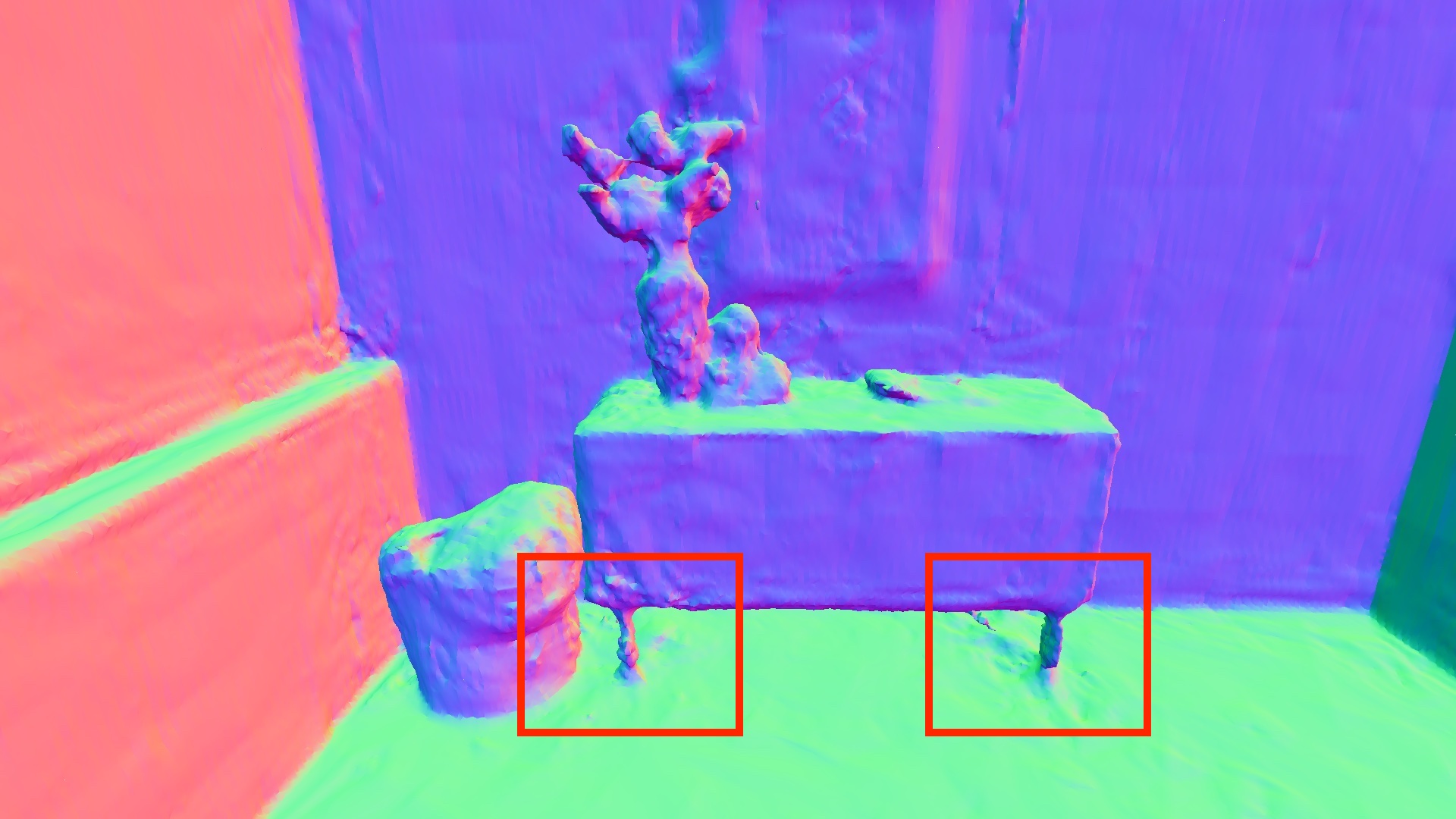}
        \caption{Ours (voxel length=0.2)}
    \end{subfigure}
    \begin{subfigure}[t]{0.33\linewidth}
        \centering
        \includegraphics[width=\linewidth]{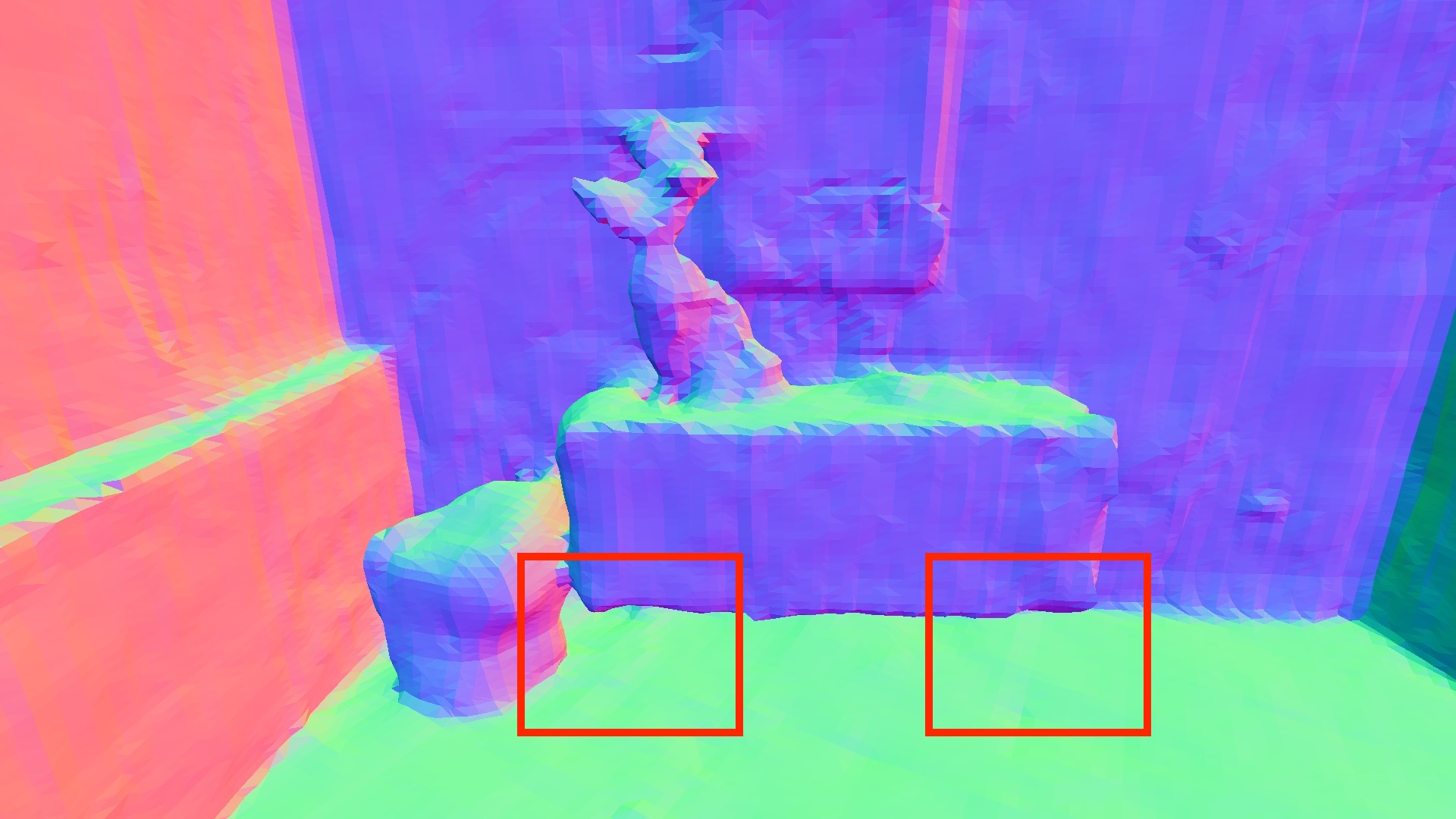}
        \caption{NICE-SLAM (voxel length=0.16)}
    \end{subfigure}
    \caption{Surface reconstruction details comparison. Our system is able to recover thin structures, such as chair legs and flowers, from the raw scans, albeit using a slightly larger voxel.}
    \label{fig:details}
\vspace{-1em}
\end{figure*}

\textbf{Key-frame selection.} In online continuous learning, keyframe selection is the key to ensuring long-term map consistency and preventing catastrophic forgetting~\cite{sucar:2021:imap}. Unlike previous works where key-frames are only inserted based on heuristically chosen metrics~\cite{sucar:2021:imap} or at a fixed interval~\cite{zhu:2021:niceslam}, our explicit voxel structure allows us to determine when to insert key-frames by performing an intersection test. Specifically, each successfully tracked frame is tested against the existing map to find the number of voxels $N_c$ that would be allocated if we are to choose it as a new keyframe. We insert a new keyframe if the ratio $p_{kf}=N_c/N_o$ is larger than a threshold, where $N_o$ is the number of currently observed voxels.

This simple strategy is adequate for exploratory movements as new voxels are constantly being allocated. However, for loopy camera motions, especially trajectories with long-term loops, there is a risk that we may never allocate new keyframes as we keep looking at an existing scene model. This will result in part of the model being completely missing or not having enough multi-view constraints. To solve this problem, we also enforce a maximum interval between adjacent frames, \textit{i.e.}, we will create a new keyframe if have not done so for the past $N$ frames. This key-frame selection strategy is simple yet effective at creating consistent scene maps.

\textbf{Joint mapping and pose update.}
Our mapping subroutine accepts tracked RGB-D frames and fuses them into the existing scene map by jointly optimizing the scene geometry and camera poses. It is noted by~\cite{sucar:2021:imap} that online incremental learning is prone to network forgetting. Therefore we use a similar method to jointly optimize the scene network and feature embeddings. For each frame, we randomly select $N_{kf}$ keyframes. These keyframes, including the recently tracked frame, can be seen as an optimization window akin to the sliding window approach employed in traditional SLAM systems~\cite{strasdat:2011:doublewindow}.

Similar to our tracking process, for each frame in the actively sampled optimization window, we randomly sample $N_m$ rays. These rays are transformed into the world coordinate system with estimated frame poses. Then we sample points within our sparse voxels and then render a set of pixels from the sample points and calculate related loss functions, using the method described in~\autoref{sec:render}. 

\subsection{Dynamic Voxel Management}
\label{sec:allocate}

Contrary to existing approaches where the full extent of the scene is encoded~\cite{sucar:2021:imap}, we are only interested in reconstructing surfaces that have observations.
Therefore, on-the-fly voxel allocation and searching are of most importance to us. For this reason, we adopt an octree structure to divide the whole scene into mutually exclusive axis-aligned voxels where we consider the voxel as the basic scene unit and the leaf node of the scene octree. An example octree is shown in~\autoref{fig:octree}. We make our system usable in unexplored areas by dynamically allocating new voxels when new observations are made.

Specifically, we initially set the leaf nodes corresponding to the unobserved scene area to empty, when a new frame is successfully tracked, we back-project its depth map into 3D points, these points are then transformed by the estimated camera pose. We then allocate new voxels for any point that does not fall into an existing voxel. Since this process needs to be applied for every point, which could have tens of thousands based on the resolution of the input images, we use an octree structure to store voxels and feature embeddings to enable fast voxel allocation and retrieval. 

\textbf{Morton coding}
Inspired by traditional volumetric SLAM systems~\cite{vespa:2018:supereight}, we choose to encode voxel coordinates as Morton codes. Morton codes are generated by interleaving the bits from each coordinate into a unique number. A 2D example of Morton coding is given in~\autoref{fig:morton}. Given the 3D coordinate $(x,y,z)$ of a voxel, we can quickly find its position in the octree by traversing through its Morton code. It is also possible to recover the encoded coordinates by applying a decoding operation. The neighbors are also identifiable by shifting the appropriate bits of the code, which is beneficial for quickly finding shared embedding vectors.

\begin{table*}
  \caption{Trajectory estimation results on the Replica dataset. Our method obtained better results on all sequences.}
  \label{tab:replica_ate}
  \scriptsize%
	\centering%
  \begin{tabu}{%
	l%
	l%
	*{2}{c}%
	*{2}{c}%
	*{2}{c}%
	*{2}{c}%
	*{2}{c}%
	*{2}{c}%
	*{2}{c}%
	*{2}{c}%
	*{2}{c}%
	}
  \toprule
  Methods & Metric & Room-0 & Room-1 & Room-2 & Office-0 & Office-1 & Office-2 & Office-3 & Office-4 & Avg. \\
  \midrule 
  \multirow{3}{*}{iMap*~\cite{sucar:2021:imap}} & \textbf{RMSE}[m]$\downarrow$ & 0.7005 & 0.0453 &0.0220 & 0.0232 & 0.0174 & 0.0487 & 0.5840 & 0.0262 & 0.1834 \\
  & \textbf{mean}[m]$\downarrow$ & 0.5891 & 0.0395 & 0.0195 & 0.01652 & 0.0155 & 0.0319  & 0.5488  & 0.0215  & 0.1603 \\
  & \textbf{median}[m]$\downarrow$ & 0.4478 & 0.0335 & 0.0173 & 0.0135 & 0.0137 & 0.0235  & 0.4756  & 0.0186  & 0.1304 \\
  \midrule
  \multirow{3}{*}{NICE-SLAM~\cite{zhu:2021:niceslam}} & \textbf{RMSE}[m]$\downarrow$ & 0.0169 & 0.0204 & 0.01554 & 0.0099 & 0.0090 & 0.0139 & 0.0397 & 0.0308 & 0.0195 \\
  & \textbf{mean}[m]$\downarrow$ & 0.0150 & 0.0180  & 0.0118  & 0.0086 & 0.0081 & 0.0120  & 0.0205 & 0.0209 & 0.0144 \\
  & \textbf{median}[m]$\downarrow$ & 0.0138 & 0.0167 & 0.0098  & 0.0076  & 0.0074  & 0.0109  & 0.0128  & 0.0153  & 0.0118 \\
  \midrule
  \multirow{3}{*}{Ours} & \textbf{RMSE}[m]$\downarrow$ & \textbf{0.0040} & \textbf{0.0054} & \textbf{0.0054} & \textbf{0.0050} & \textbf{0.0046} & \textbf{0.0075} & \textbf{0.0050} & \textbf{0.0060} & \textbf{0.0054} \\
  & \textbf{mean}[m]$\downarrow$ & \textbf{0.0036} & \textbf{0.0043} & \textbf{0.0041} & \textbf{0.0040} & \textbf{0.0043} & \textbf{0.0058} & \textbf{0.0045} & \textbf{0.0055} & \textbf{0.0045} \\
  & \textbf{median}[m]$\downarrow$ & \textbf{0.0033} & \textbf{0.0038} & \textbf{0.0036} & \textbf{0.0035} & \textbf{0.0041} & \textbf{0.0048} & \textbf{0.0042} & \textbf{0.0053} & \textbf{0.0041} \\
  \bottomrule
  \end{tabu}%
\end{table*}

\begin{figure*}
    \centering
    \begin{subfigure}[t]{0.24\linewidth}
        \centering
        \includegraphics[width=\textwidth]{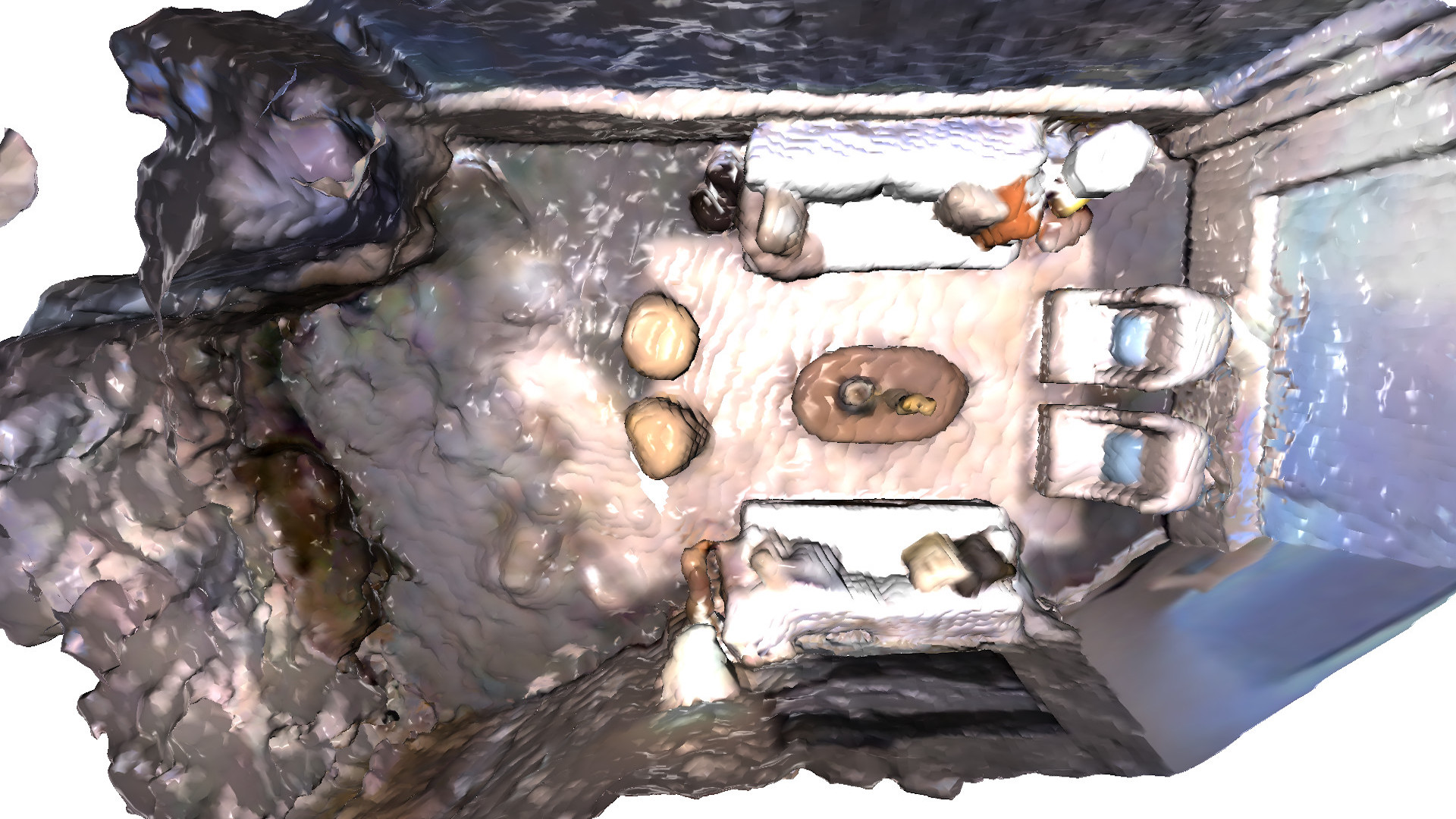}
    \end{subfigure}
    \begin{subfigure}[t]{0.24\linewidth}
        \centering
        \includegraphics[width=\textwidth]{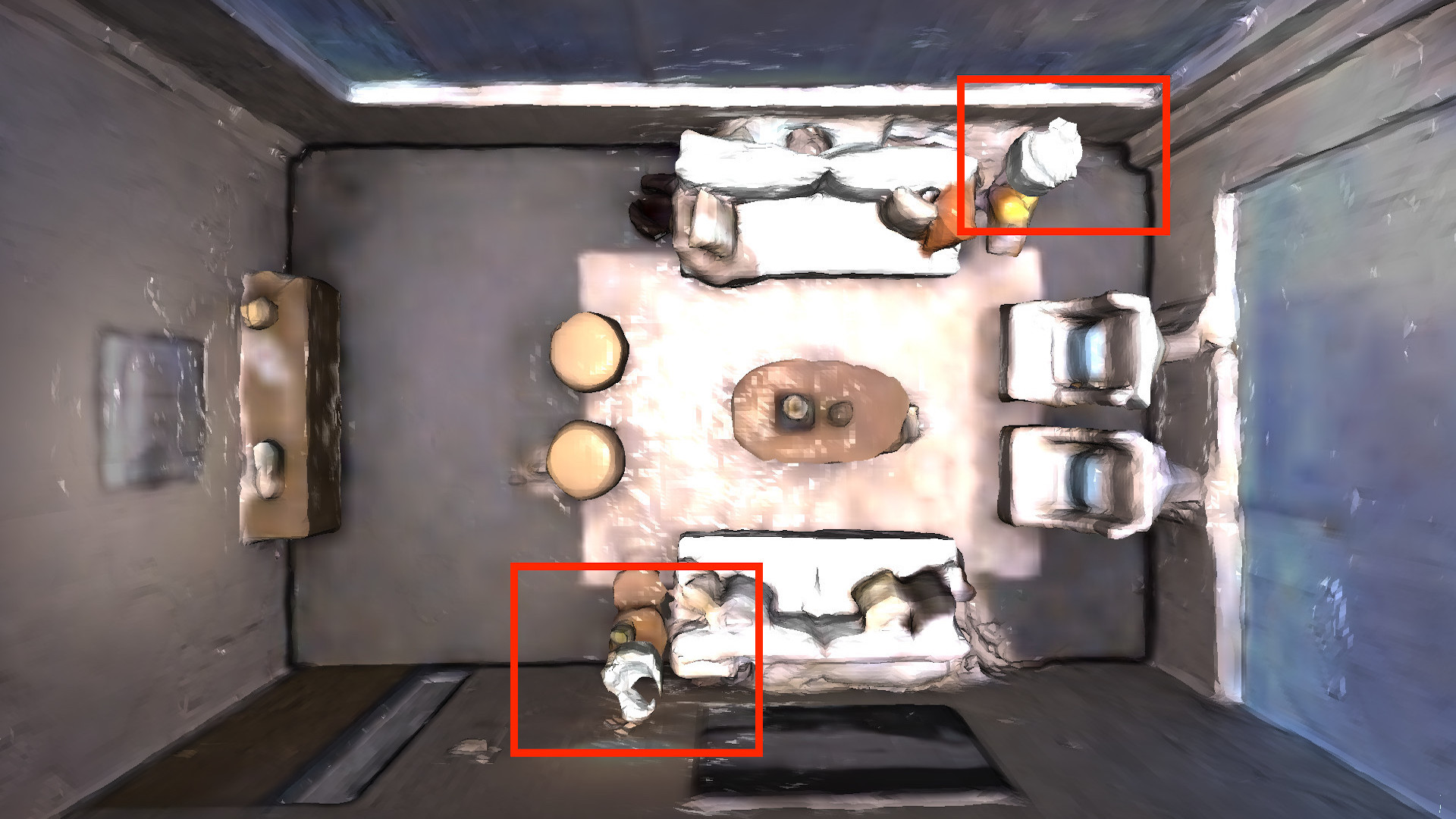}
    \end{subfigure}
    \begin{subfigure}[t]{0.24\linewidth}
        \centering
        \includegraphics[width=\textwidth]{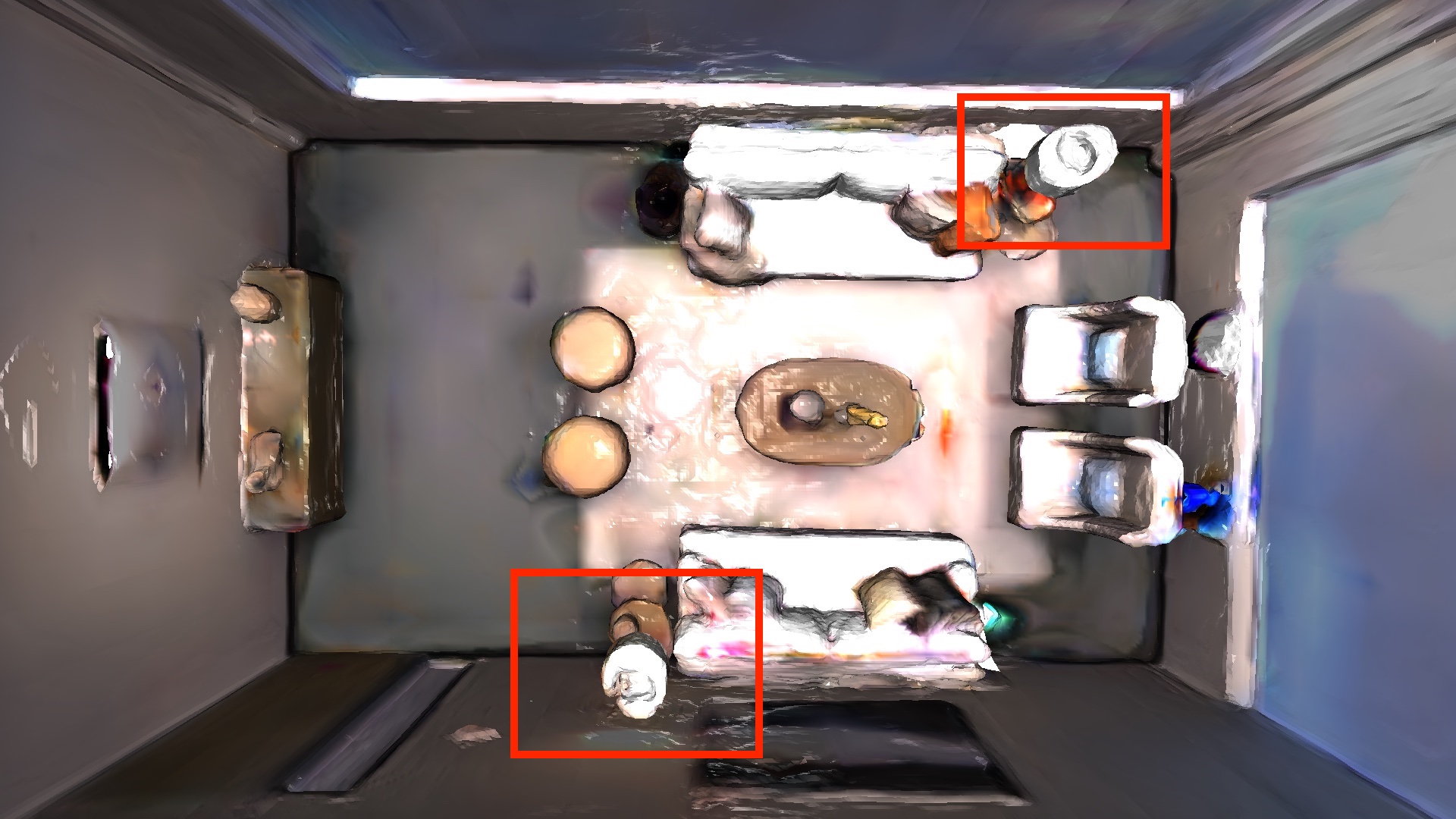}
    \end{subfigure}
    \begin{subfigure}[t]{0.24\linewidth}
        \centering
        \includegraphics[width=\textwidth]{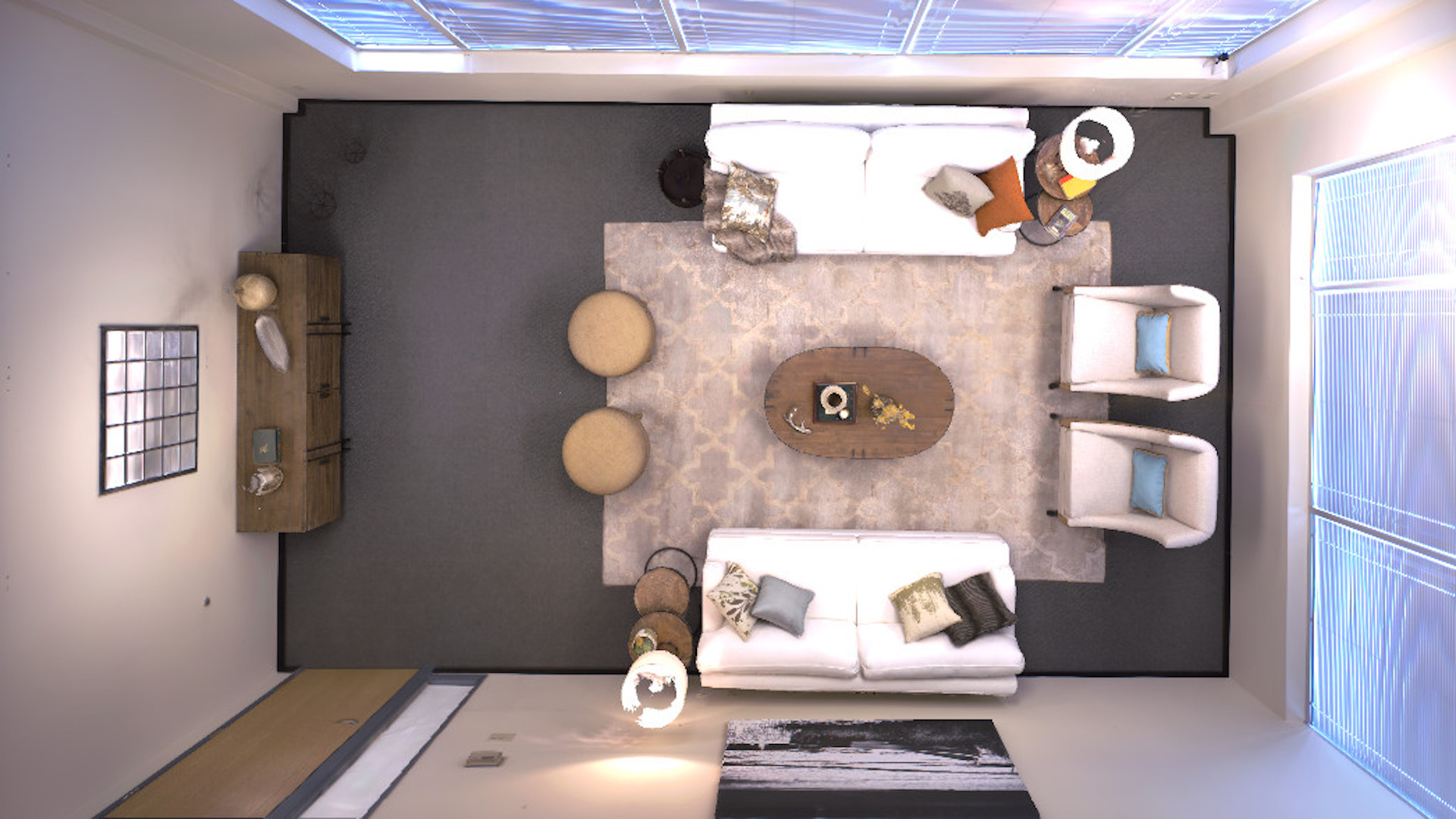}
    \end{subfigure}
    \hfill
    \begin{subfigure}[t]{0.24\linewidth}
        \centering
        \includegraphics[width=\textwidth]{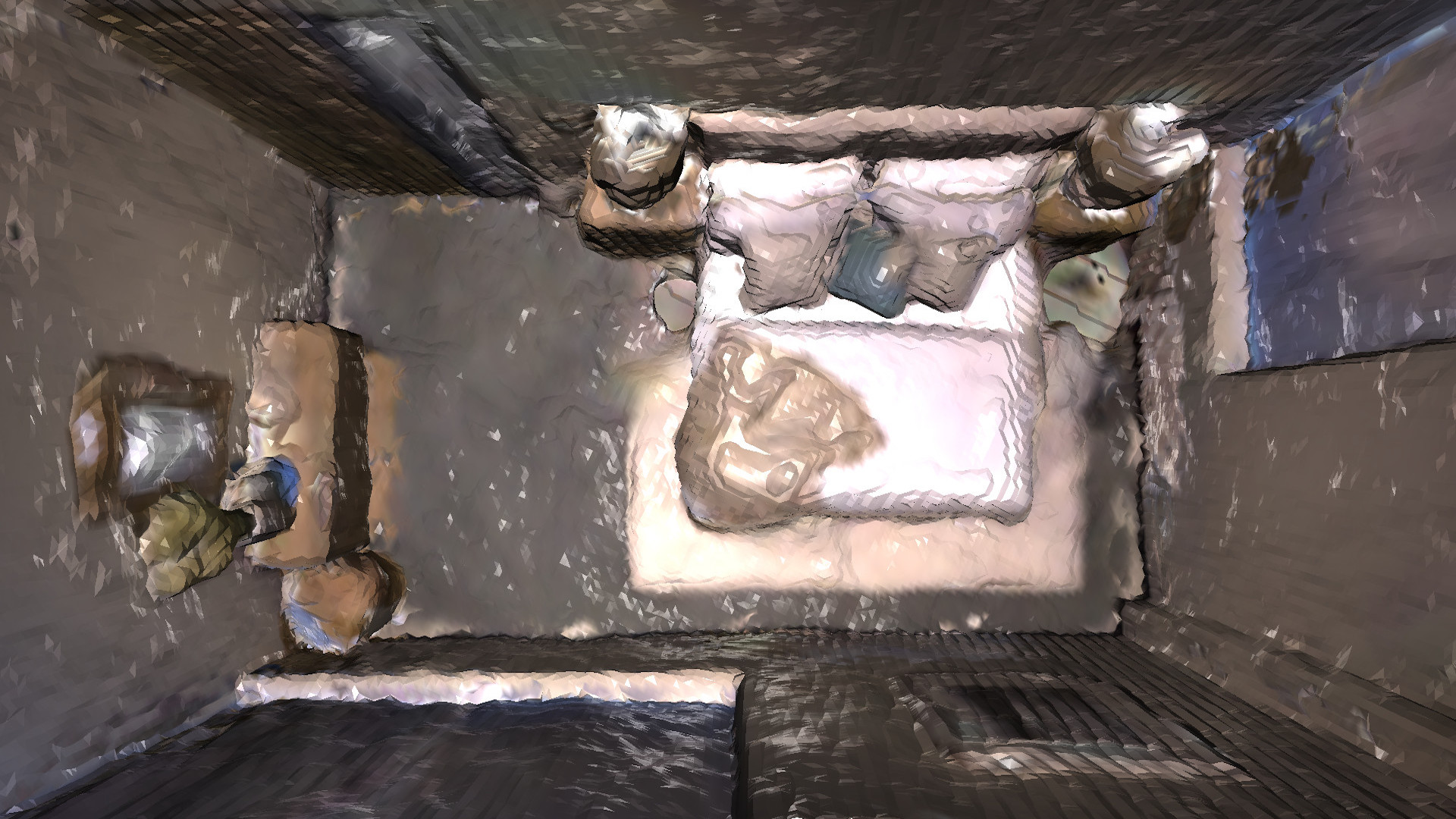}
    \end{subfigure}
    \begin{subfigure}[t]{0.24\linewidth}
        \centering
        \includegraphics[width=\textwidth]{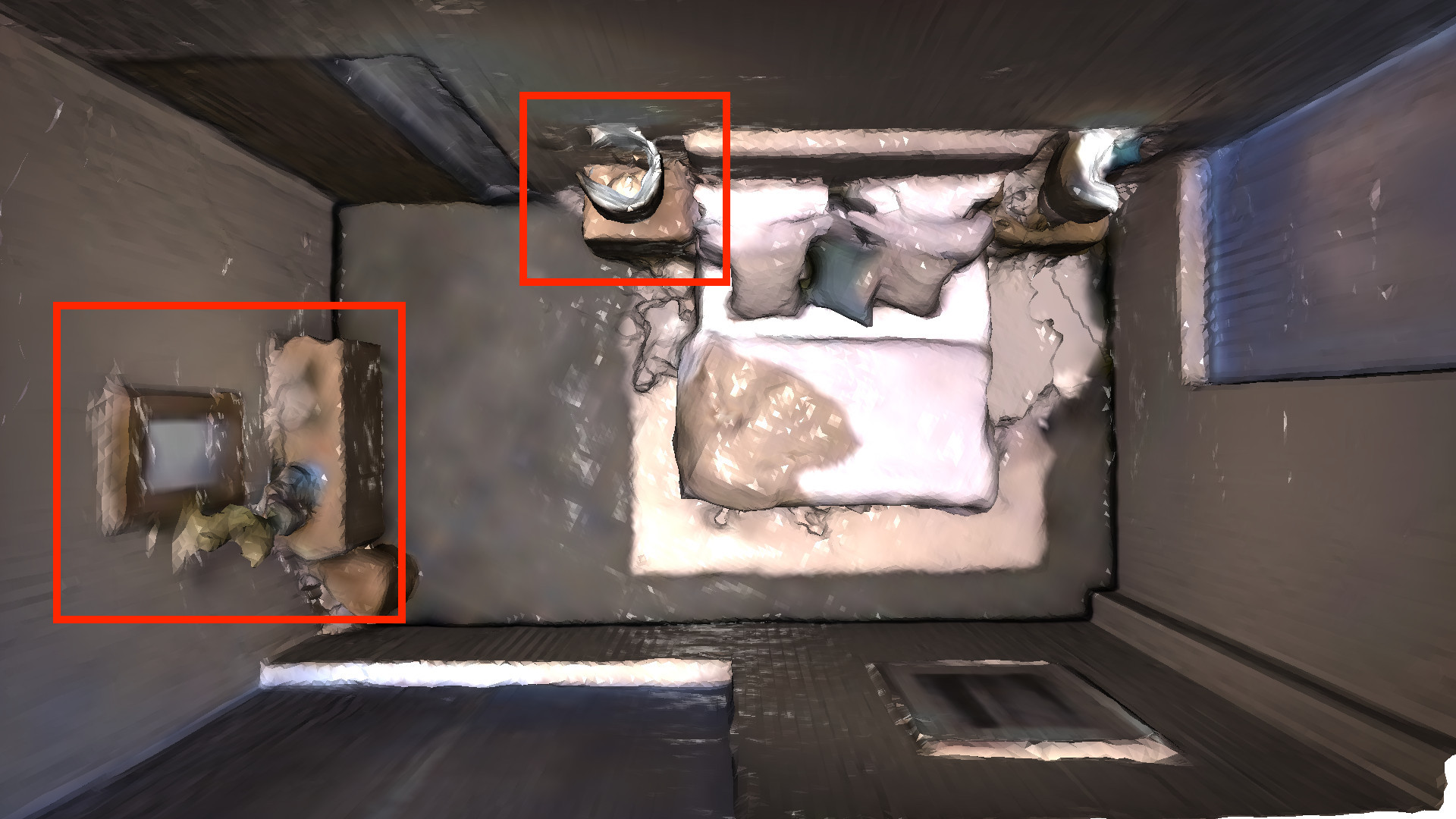}
    \end{subfigure}
    \begin{subfigure}[t]{0.24\linewidth}
        \centering
        \includegraphics[width=\textwidth]{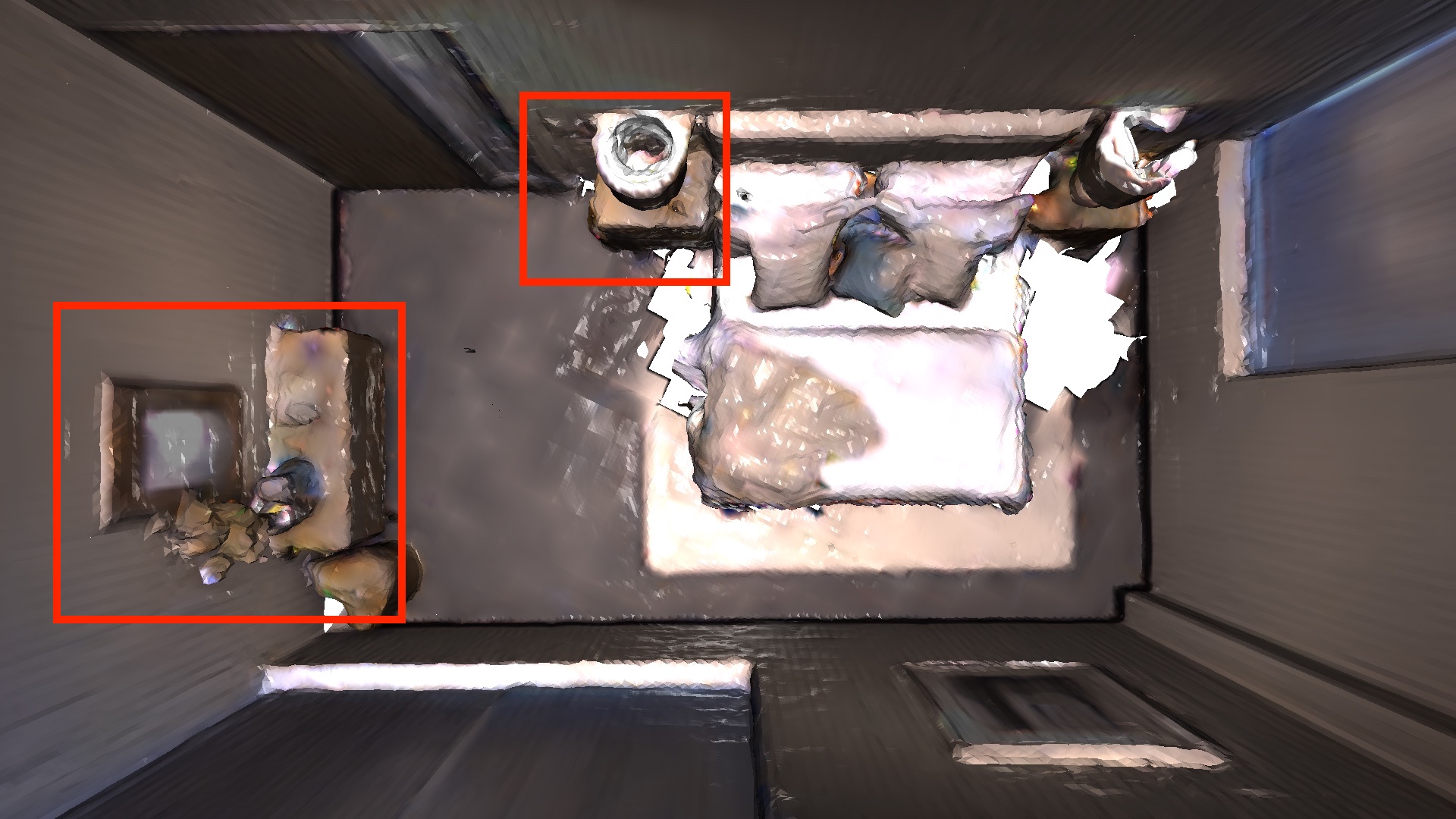}
    \end{subfigure}
    \begin{subfigure}[t]{0.24\linewidth}
        \centering
        \includegraphics[width=\textwidth]{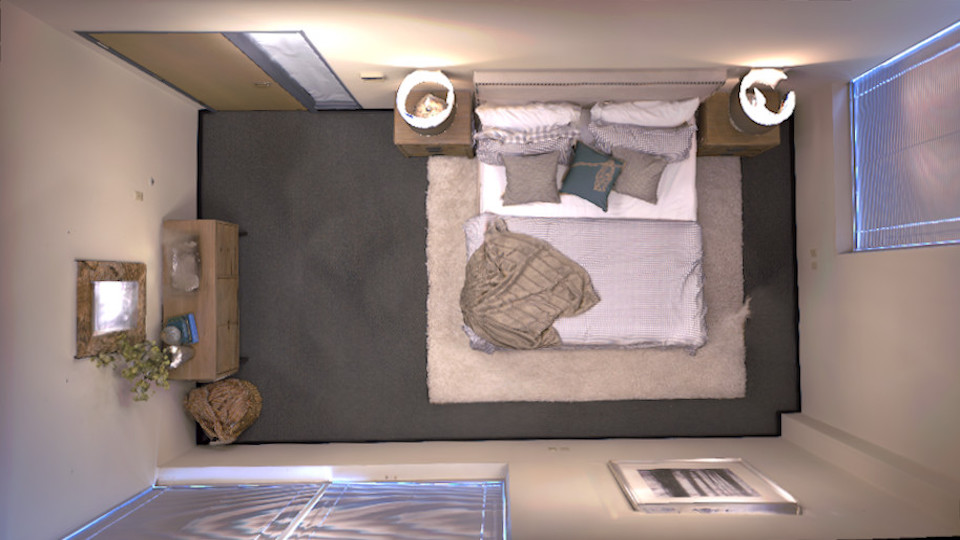}
    \end{subfigure}
    \hfill
    \begin{subfigure}[t]{0.24\linewidth}
        \centering
        \includegraphics[width=\textwidth]{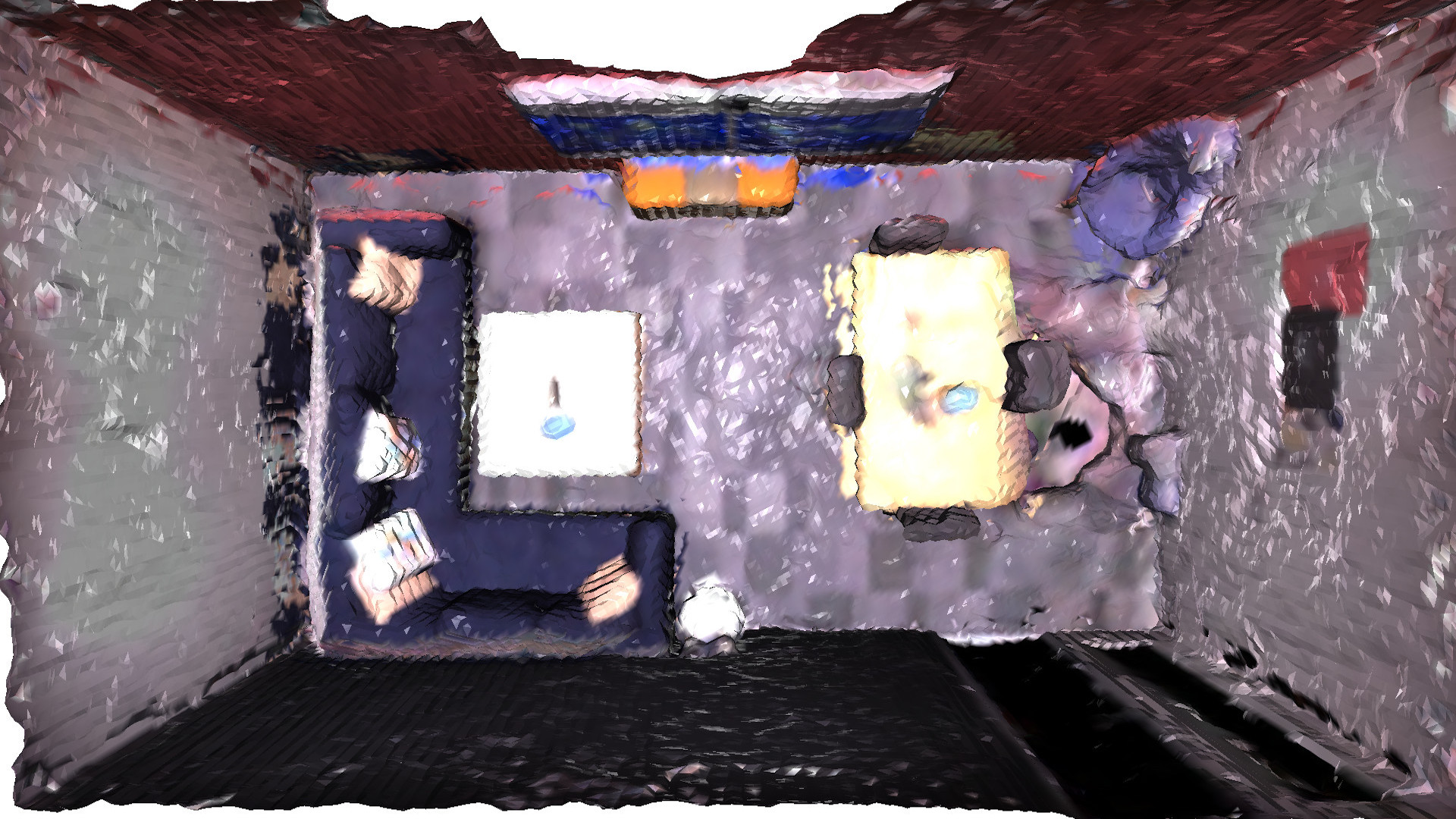}
    \end{subfigure}
    \begin{subfigure}[t]{0.24\linewidth}
        \centering
        \includegraphics[width=\textwidth]{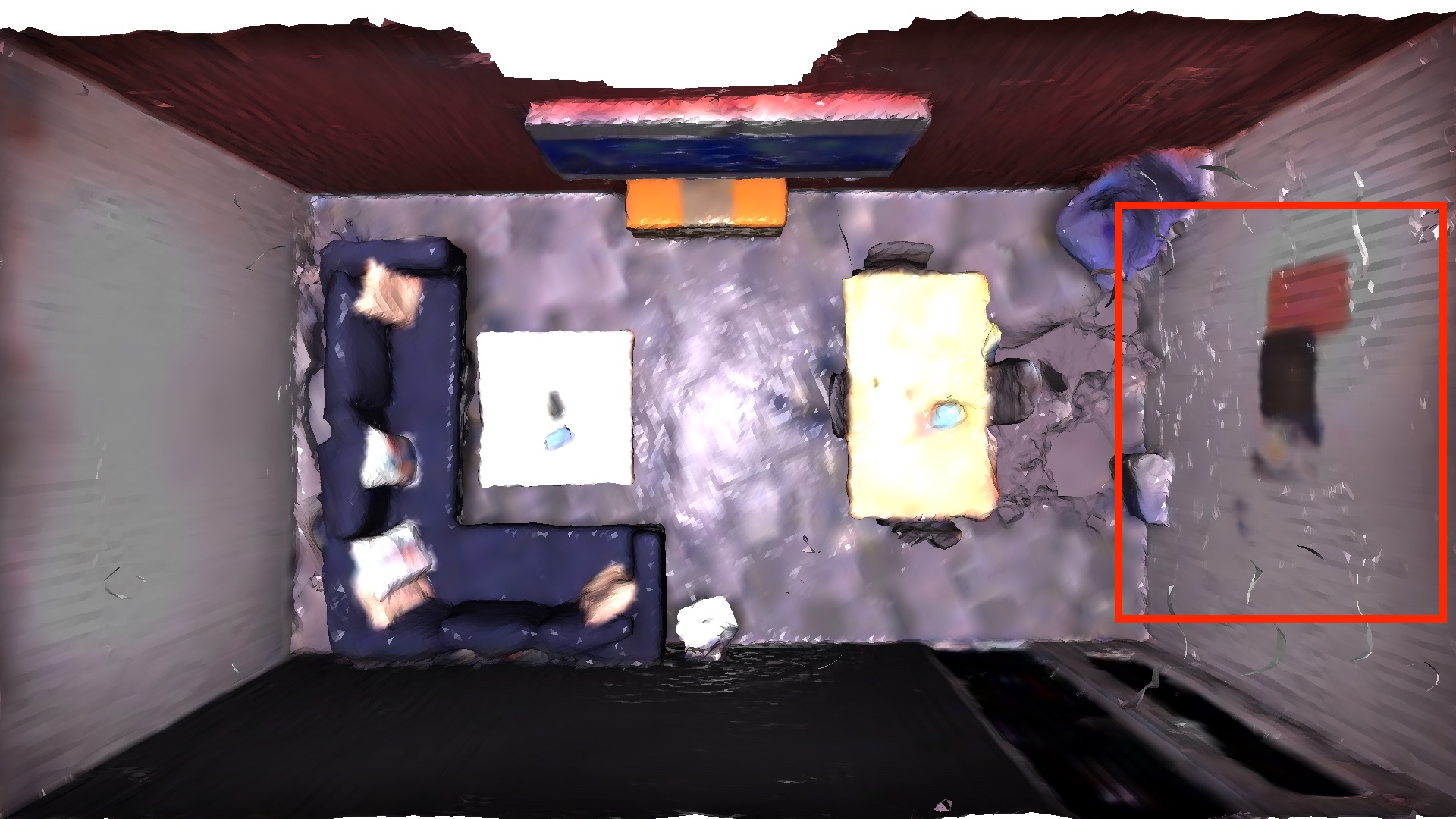}
    \end{subfigure}
    \begin{subfigure}[t]{0.24\linewidth}
        \centering
        \includegraphics[width=\textwidth]{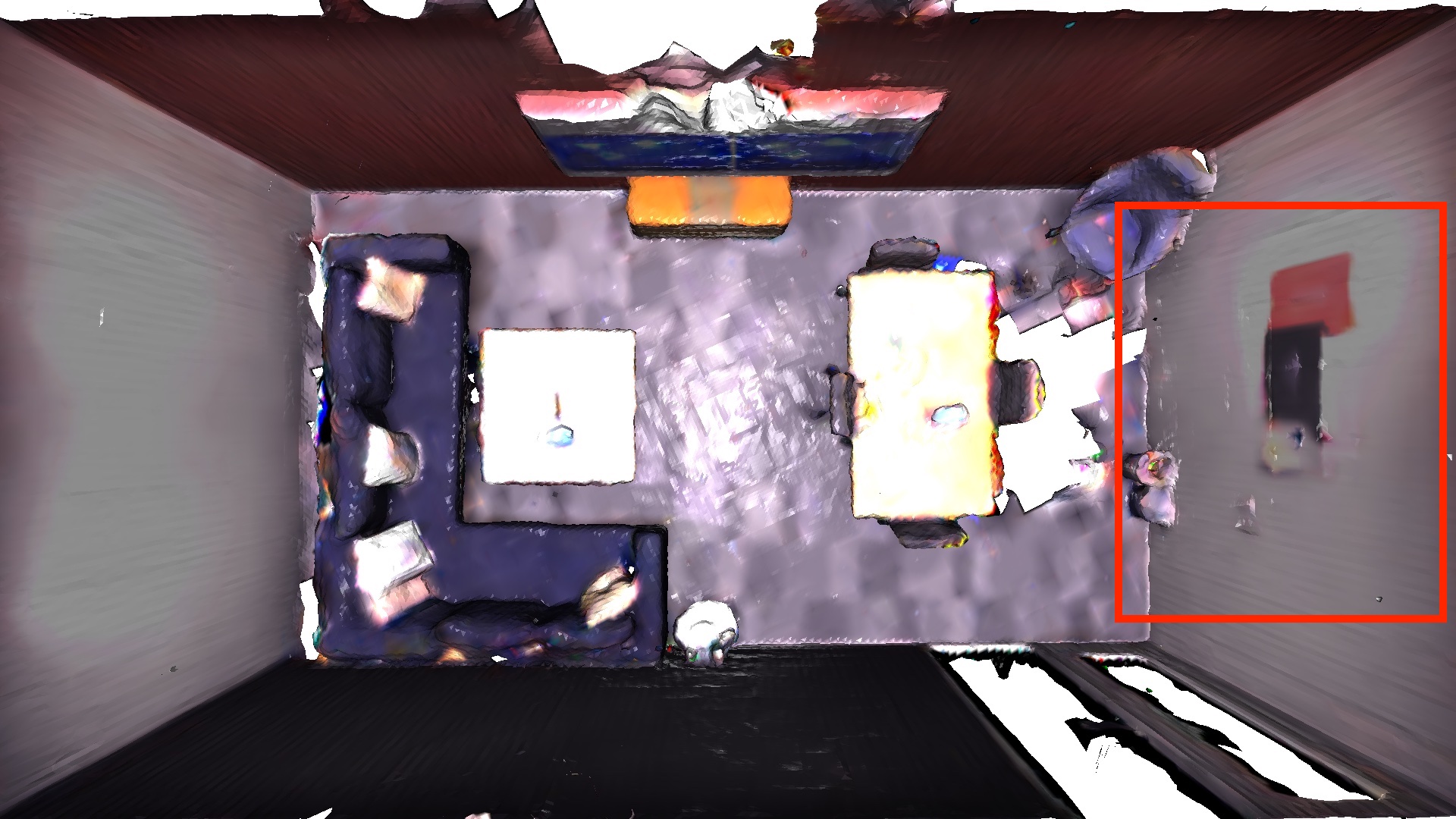}
    \end{subfigure}
    \begin{subfigure}[t]{0.24\linewidth}
        \centering
        \includegraphics[width=\textwidth]{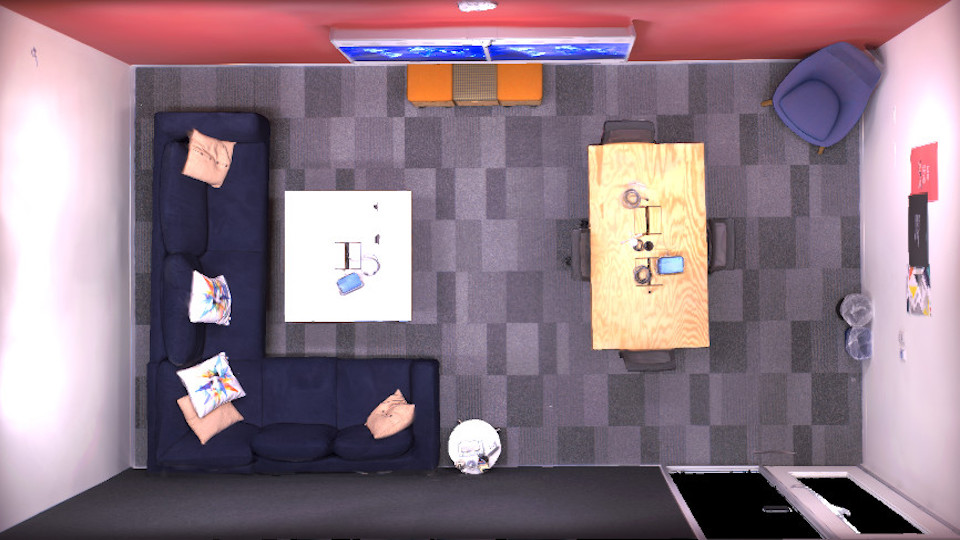}
    \end{subfigure}
    \hfill
    \begin{subfigure}[t]{0.24\linewidth}
        \centering
        \includegraphics[width=\textwidth]{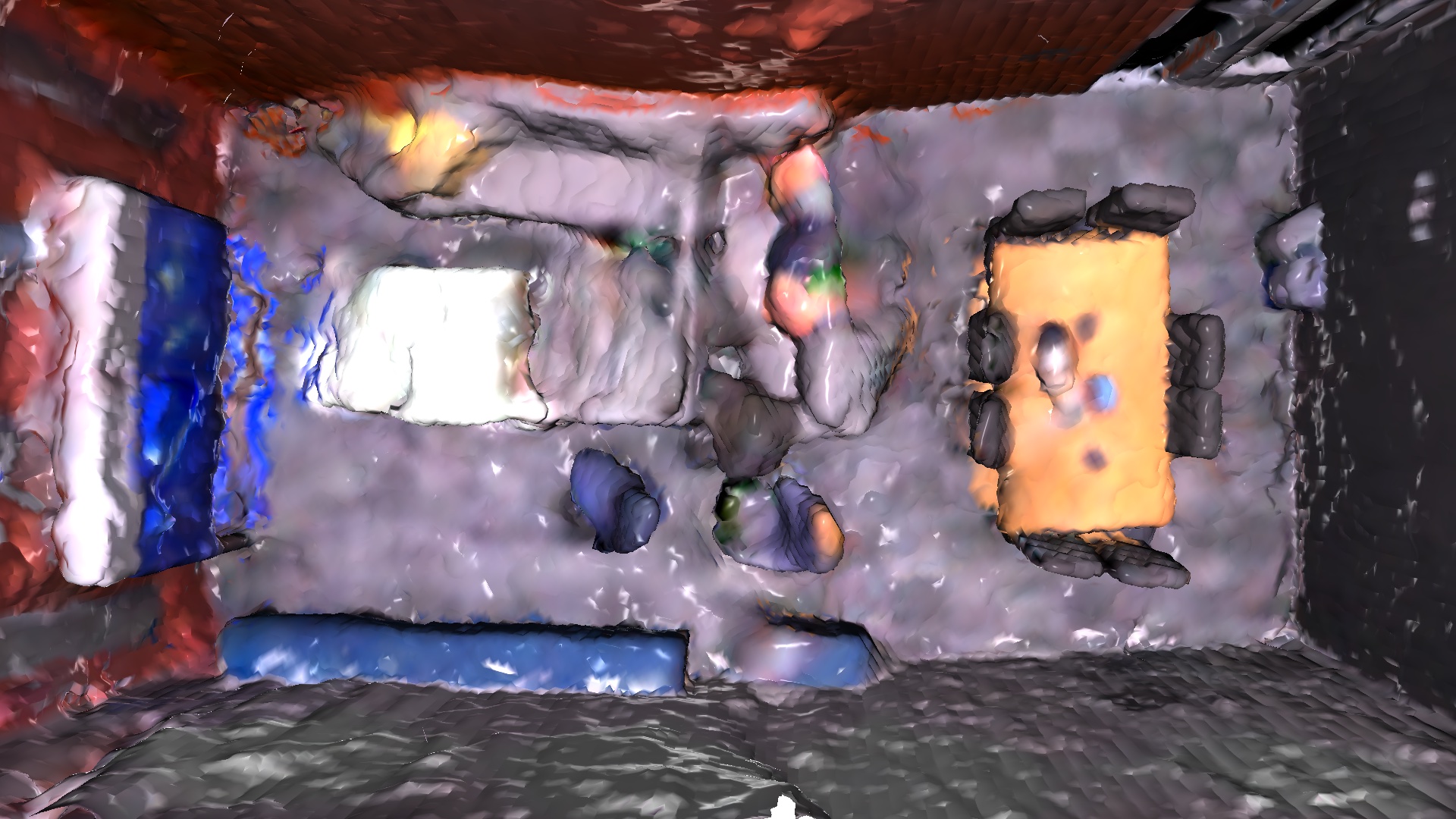}
        \caption{iMap$^{*}$}
    \end{subfigure}
    \begin{subfigure}[t]{0.24\linewidth}
        \centering
        \includegraphics[width=\textwidth]{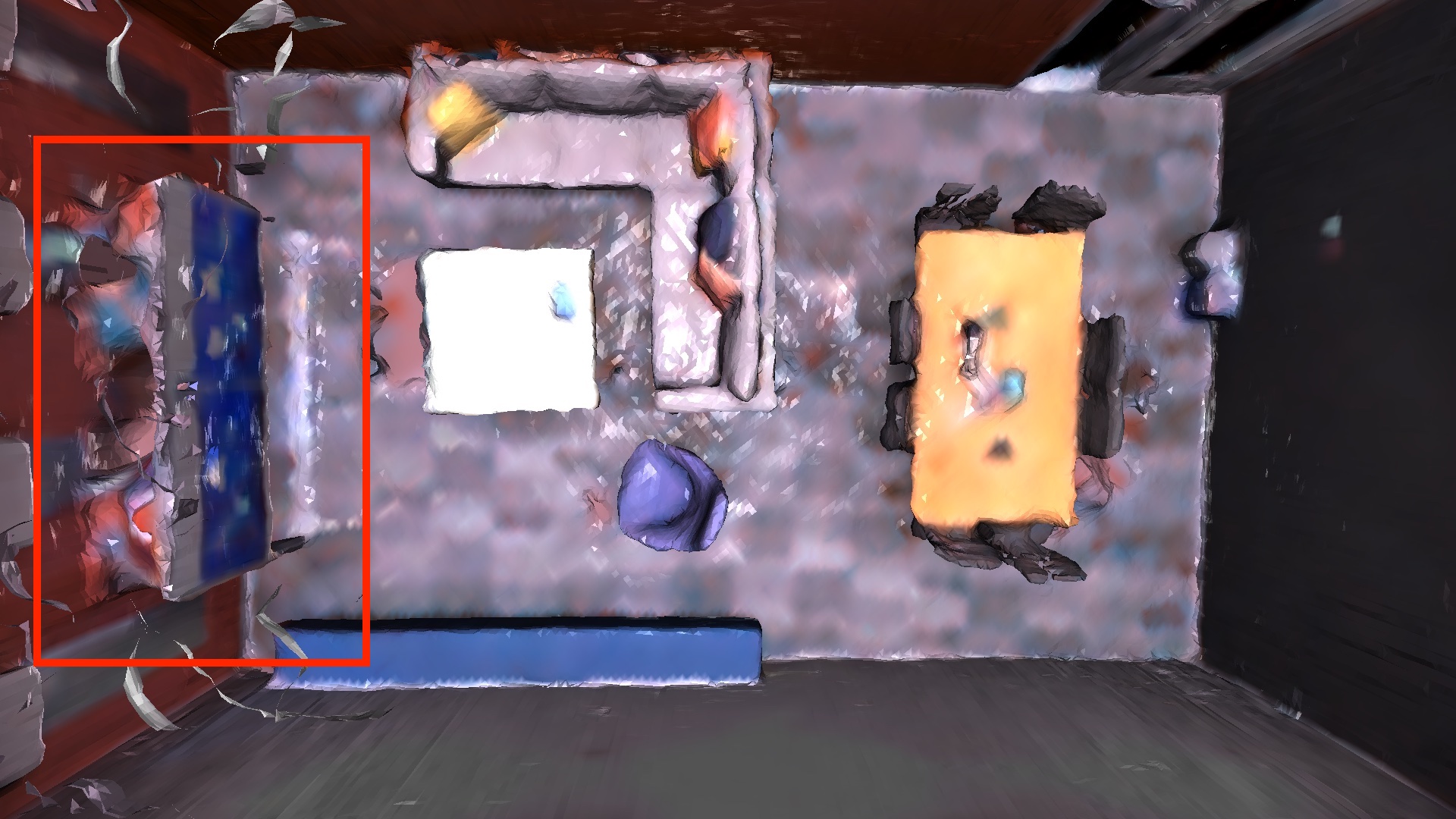}
        \caption{NICE-SLAM}
    \end{subfigure}
    \begin{subfigure}[t]{0.24\linewidth}
        \centering
        \includegraphics[width=\textwidth]{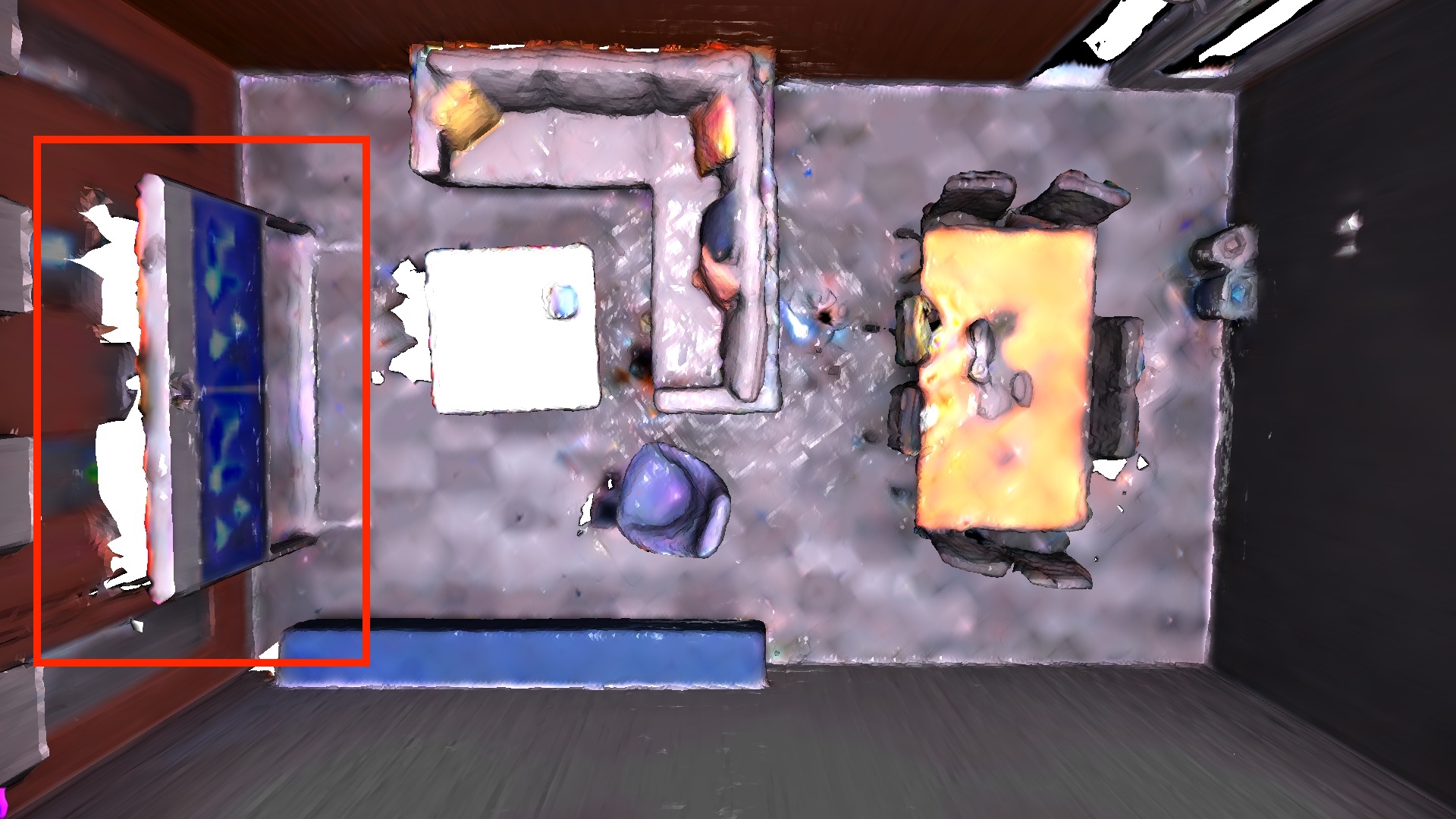}
        \caption{Ours}
    \end{subfigure}
    \begin{subfigure}[t]{0.24\linewidth}
        \centering
        \includegraphics[width=\textwidth]{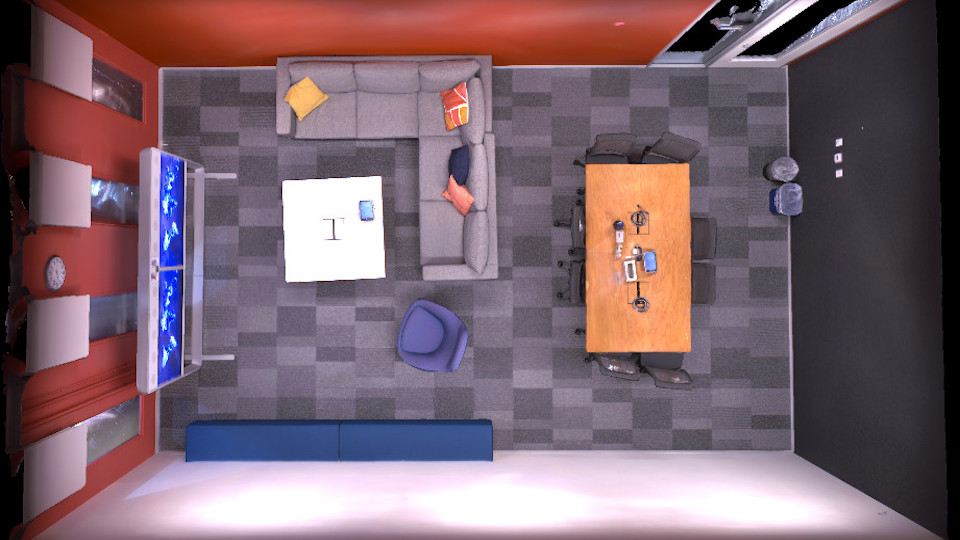}
        \caption{Ground truth}
    \end{subfigure}
    \caption{Qualitative reconstruction results on the Replica dataset. From left to right, we show the results of scene reconstruction of different methods (iMAP$^{*}$, NICE-SLAM, our method, and ground truth). It can be clearly seen that our reconstruction results are much better than iMAP$^{*}$. To better show the difference in reconstruction between NICE-SLAM and our method, we use red boxes in the figures to indicate the improvements over NICE-SLAM.}
    \label{fig:replica}
\end{figure*}

\begin{table*}
  \caption{Reconstruction Results of 8 Scenes in the Replica dataset. Compared with iMAP and NICE-SLAM, our approach yields better results consistently.}
  \label{tab:replica}
  \scriptsize%
	\centering%
  \begin{tabu}{%
	l%
	l%
	*{2}{c}%
	*{2}{c}%
	*{2}{c}%
	*{2}{c}%
	*{2}{c}%
	*{2}{c}%
	*{2}{c}%
	*{2}{c}%
	*{2}{c}%
	}
  \toprule
  Methods & Metric & Room-0 & Room-1 & Room-2 & Office-0 & Office-1 & Office-2 & Office-3 & Office-4 & Avg. \\
 \midrule 
  \multirow{3}{*}{iMap~\cite{sucar:2021:imap}} & \textbf{Acc.}[cm]$\downarrow$ & 3.58 & 3.69 & 4.68 & 5.87 & 3.71 & 4.81 & 4.27 & 4.83 & 4.43 \\
                          & \textbf{Comp.}[cm]$\downarrow$ & 5.06 & 4.87 & 5.51 & 6.11 & 5.26 & 5.65 & 5.45 & 6.59 & 5.56 \\
                          & \textbf{Comp. Ratio}[$<5$cm \%]$\uparrow$ & 83.91 & 83.45 & 75.53 & 77.71 & 79.64 & 77.22 & 77.34 & 77.63 & 79.06\\
  \midrule
  \multirow{3}{*}{NICE-SLAM~\cite{zhu:2021:niceslam}} & \textbf{Acc.}[cm]$\downarrow$ & 3.53 & 3.60 & \textbf{3.03} & 5.56 & 3.35 & 4.71 & 3.84 & 3.35 & 3.87 \\
                          & \textbf{Comp.}[cm]$\downarrow$ & 3.40 &  3.62 & 3.27 & 4.55 & 4.03 & 3.94 & 3.99 & 4.15 & 3.87 \\
                          & \textbf{Comp. Ratio}[$<5$cm \%]$\uparrow$ & 86.05 & 80.75 & 87.23 & 79.34 & 82.13 & 80.35 & 80.55 & 82.88 & 82.41 \\
  \midrule
  \multirow{3}{*}{Ours} & \textbf{Acc.}[cm]$\downarrow$ & \textbf{2.41} & \textbf{1.62} & 3.11 & \textbf{1.74} & \textbf{1.69} & \textbf{2.23} & \textbf{2.84} & \textbf{3.31} & \textbf{2.37} \\
                          & \textbf{Comp.}[cm]$\downarrow$ & \textbf{2.60} & \textbf{2.23} & \textbf{1.93} & \textbf{1.39} & \textbf{1.80} & \textbf{2.71} & \textbf{2.69} & \textbf{2.88} & \textbf{2.28} \\
                          & \textbf{Comp. Ratio}[$<5$cm \%]$\uparrow$ & \textbf{92.87} & \textbf{93.48} & \textbf{94.34} & \textbf{97.21} & \textbf{93.76} & \textbf{90.98} & \textbf{90.73} & \textbf{89.48} & \textbf{92.86} \\
  \bottomrule
  \end{tabu}%
\end{table*}

\begin{figure*}
    \centering
    \includegraphics[width=\linewidth]{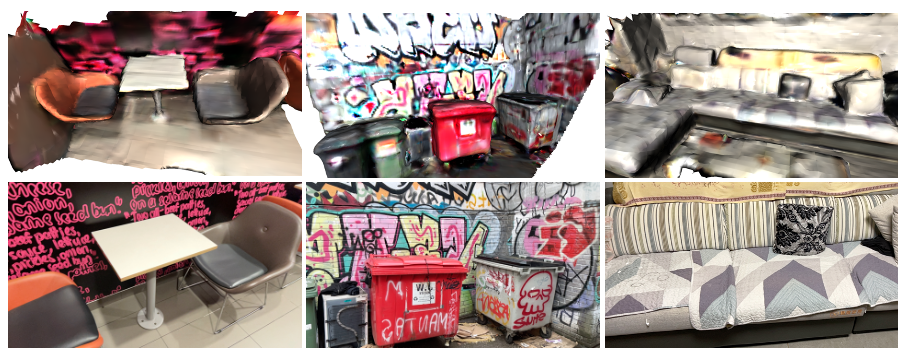}
    \vspace{-2em}
    \caption{Reconstruction of real-world RGB-D sequences captured with a handheld device. The images were taken by an iPhone 13 Pro. We get depth images directly from the onboard lidar sensor. We show (top) reconstructions and (bottom) an example of the original input image. }
    \label{fig:ios}
\vspace{-1em}
\end{figure*}

\section{Experiments}
\label{sec:exp}

\subsection{Experimental Setup}
\label{subsec:setup}

\textbf{Datasets}: In our experiments we use $3$ different datasets: (1) Replica dataset~\cite{julian:2019:replica} which contains $18$ different scenes captured by a camera rig. Based on Replica dataset, \cite{sucar:2021:imap} further synthesizes the RGB-D sequences, which are used in our experiments. (2) ScanNet dataset~\cite{dai:2017:scannet} which contains more than $1000$ captured RGB-D sequences and ground truth poses estimated from a SLAM system~\cite{dai:2017:bundlefusion}. (3) Several indoor and outdoor RGB-D sequences that were captured by iOS devices equipped with range sensors such as iPhone 13 Pro and iPad Pro. These datasets cover a wide range of applications and scenarios, therefore are well suited to study our proposed system.

\textbf{Evaluation Metrics}: We use several different metrics to measure the performance of our system as well as other competing methods. For reconstruction quality, we measure accuracy and completion, this is in line with previous works~\cite{sucar:2021:imap, zhu:2021:niceslam}. Mesh accuracy (Acc.) is defined as the un-directional Chamfer distance from the reconstructed mesh to the ground-truth. Completion (Comp.) is similarly defined as the distance the other way around. We also measure the completion ratio, which is the percentage of the reconstructed points whose distance to the ground truth mesh is smaller than $5cm$. We show the formulation of Chamfer distance in ~\autoref{eqn:chamfer}, where $p \in P$ and $q \in Q$ are two point sets sampled from the reconstructed and ground truth meshes: 

\begin{equation}
\label{eqn:chamfer}
\begin{split}
    d_{Chamfer} = |P|^{-1} \sum_{(p,q) \in \Lambda_{P,Q}} \Vert p-q \Vert^2, \\
    \Lambda_{Q,P}^* = \{(p, argmin_{q} \Vert p - q \Vert)\}.
\end{split}
\end{equation}

To benchmark pose estimation, we adopt the commonly used absolute trajectory error (ATE) using the scripts provided by~\cite{sturm:2012:tumrgbd}. ATE is calculated as the absolute translational difference between the estimated and ground truth poses.

\textbf{Implementation Details}: We illustrate our network architecture in~\autoref{fig:arch}. Our decoder is implemented as an MLP consisting of several fully-connected layers (FC) and skip connections. The input to our network is a $16$-$D$ feature embedding. The features are generally processed by $2$-$4$ FC layers that each has $256$ hidden units. The SDF head outputs a scalar SDF value $s$ and a $128$-$D$ hidden vector. The color head has two FC layers with $256$ hidden units each. We apply sigmoid to generate RGB values in the range $[0, 1]$. The step size ratio for sampling voxel points is generally set to $0.05$-$0.1$. For all scenes, we use a voxel size of $0.2m$.

\subsection{Reconstruct Synthetic Scenes}

To test our system on reconstructing synthetic scenes, we use the Replica dataset. The RGB-D sequences we use are released by~\cite{sucar:2021:imap} and subsequently used in NICE-SLAM~\cite{zhu:2021:niceslam}. We compare our system qualitatively with iMap and NICE-SLAM. The results of iMap are directly obtained from their paper, while the results for NICE-SLAM is generated from their official code release. We show the qualitative evaluation results in~\autoref{fig:replica}. It can be seen that our method produces better maps than iMap, and performs on par with NICE-SLAM. 

It is worth noting that both NICE-SLAM and iMap assume densely populated surfaces, therefore they will create surfaces even in places that is not observed. Surfaces created in this way will be realistic when the gap is small, but deviates from the ground truth by a large margin when there is a huge gap. Our use of an explicit voxel map prevents this from happening, i.e., our system only hallucinates surfaces inside the visible sparse voxels, therefore we can still produce plausible hole fill-in effects, while leaving large unobserved spaces empty. By doing so, we effectively combine the best of both worlds. Although it might seem like a disadvantage at first, we argue that for real-world tasks it is often more important to know where has been observed and where has not.

Owing to the expressiveness of the signed distance representation, Our method is also able to reconstruct more detailed surfaces. This is shown in~\autoref{fig:details}. In our results, the table legs and flowers are clearly seen but missing in the results of NICE-SLAM, despite that we use slightly larger voxels and only a single voxel grid level as opposed to three levels in NICE-SLAM. 

We also quantitatively compared our system on reconstruction quality and trajectory estimation with iMap and NICE-SLAM. Please note that the results of reconstruction quality for iMap are directly obtained from its paper~\cite{sucar:2021:imap}, while the trajectory estimation experiment was performed with the iMap implementation of~\cite{zhu:2021:niceslam} since the original authors do not open source their code (denoted as iMap*). 
The results for NICE-SLAM are taken from the supplementary material of the published paper. Please note that we use the results computed without mesh culling for a fair comparison.
The results on reconstruction accuracy are listed in~\autoref{tab:replica}. The results for trajectory estimation accuracy are listed in~\autoref{tab:replica_ate}. It is clear that we surpass both systems on all metrics. We also obtained much better results on camera pose estimation with a large margin. These results further confirm our observation that our system is able to produce state-of-the-art results on synthetic datasets.

\begin{figure*}
    \centering
    \begin{subfigure}[t]{0.24\textwidth}
        \centering
        \includegraphics[width=\textwidth]{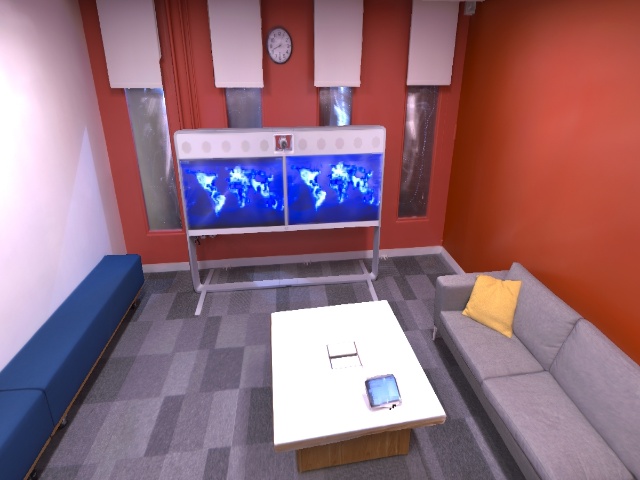}
        \caption{Ground Truth}
    \end{subfigure}
    \begin{subfigure}[t]{0.24\textwidth}
        \centering
        \includegraphics[width=\textwidth]{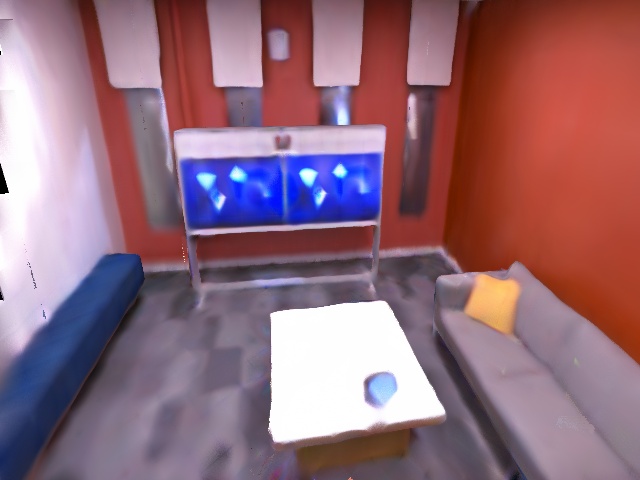}
        \caption{Rendered Image}
    \end{subfigure}
    \begin{subfigure}[t]{0.24\textwidth}
        \centering
        \includegraphics[width=\textwidth]{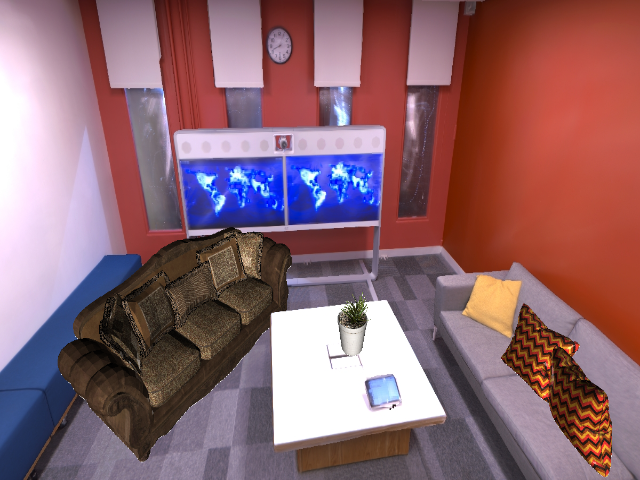}
        \caption{AR View1}
    \end{subfigure}
    \begin{subfigure}[t]{0.24\textwidth}
        \centering
        \includegraphics[width=\textwidth]{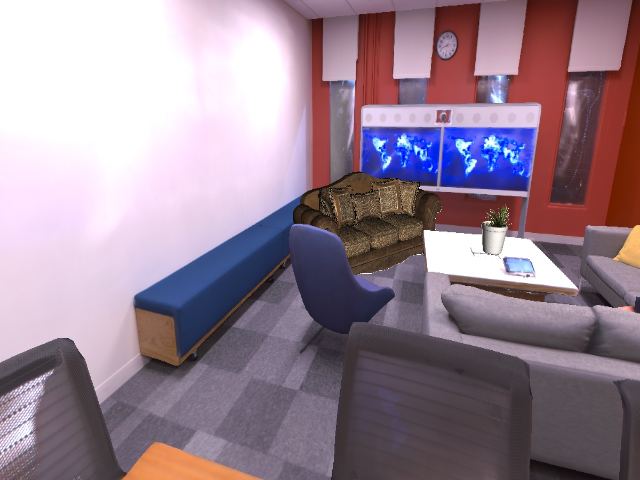}
        \caption{AR View2}
    \end{subfigure}
    \vspace{-1em}
    \caption{The ground truth image (a) in the Replica office3 dataset and its rendered image (b) with our reconstructed scene. And we place some pre-defined objects in office3, which is shown by (c) and (d) in different viewpoints. We show that we can achieve a good occlusion relationship between real and virtual objects.}
    \label{fig:ar_images}
\vspace{-1em}
\end{figure*}

\subsection{Reconstruct Real Scans}

\begin{table}
  \caption{Trajectory estimation results on the ScanNet dataset (RMSE).}
  \label{tab:scannet_ate}
  \scriptsize%
	\centering%
  \begin{tabu}{%
	l%
	*{2}{c}%
	*{2}{c}%
	*{2}{c}%
	*{2}{c}%
	*{2}{c}%
	*{2}{c}%
	}
  \toprule
  Scene ID & 0000 & 0106 & 0169 & 0181 & 0207 \\
  \midrule 
  DI-Fusion~\cite{huang:2021:difusion} & 0.6299 & 0.1850 & 0.7580 & 0.8788 & 1.0019 \\
  iMap*~\cite{sucar:2021:imap} & 0.5595 &0.1750 & 0.7051 & 0.3210 & 0.1191 \\
  NICE-SLAM~\cite{zhu:2021:niceslam} & 0.0864 & 0.0809 & 0.1028 & 0.1293 & 0.0559 \\
  Ours & \textbf{0.0839} & \textbf{0.0744} & \textbf{0.0653} & \textbf{0.1220} & \textbf{0.0557} \\
  \bottomrule
  \end{tabu}%
\vspace{-1em}
\end{table}

Unlike synthetic datasets, real scans are noisier and contain erroneous measurements. Reconstructing real scans is considered a challenging task that has not yet been solved. We benchmarked our system on selected sequences of ScanNet~\cite{dai:2017:scannet}. The selection of sequences is in line with~\cite{zhu:2021:niceslam}, and the results from DI-Fusion~\cite{huang:2021:difusion}, iMap* and NICE-SLAM are directly taken from~\cite{zhu:2021:niceslam}. The quantitative results are shown in~\autoref{tab:scannet_ate}. 
It can be seen that despite the simplicity of our design, our method still achieves better results than iMap and NICE-SLAM. We also provide geometric reconstruction with sufficient details, which provides better results in scene accuracy and completion.

We also tested our system on reconstructing outdoor RGB-D sequences captured with a handheld device. In our case, we take these images with an iPhone 13 Pro and an iPad Pro (2020). Our system is able to reconstruct the scene with reasonable accuracy without prior knowledge of the scene. We demonstrate the results in~\autoref{fig:ios}. Please note that iOS devices can only provide depth maps with very low resolution, which limits the reconstruction quality. As can be seen from the figure, our system can reconstruct the scene in various scenarios, and we believe this is a big step towards truly useful neural implicit SLAM systems.

\subsection{Time and Memory Efficiency}

We use a highly efficient multi-process implementation for the parallel tracking and mapping. Since we make a local copy of shared resources (e.g. voxels, features, and the implicit decoder, etc.) for the tracking process each time the map is updated, the probability of resource contention between the two processes is low. The performance hit of accessing the same resource can be further minimized using better engineered lock-free structures, which are not covered in our work. 

\begin{table}
\caption{Average time spent on each component.}
\label{tab:time}
  \scriptsize%
	\centering%
  \begin{tabu}{%
	l%
	*{7}{c}%
	}
  \toprule
  Components & Measured time \\
  \midrule
    Tracking & 12 ms \\
    Mapping & 55 ms \\
  \midrule
    Voxel allocation & 0.1 ms \\
    Ray-voxel intersection test & 0.9 ms \\
    Point sampling & 1 ms \\
    Volume rendering & 4 ms \\
    Back-propagation & 6 ms \\
  \bottomrule
  \end{tabu}%
\vspace{-1em}
\end{table}

To study the impact on running time for our sparse voxel-based sampling and rendering method, we profile our system on the synthetic Replica dataset. We measure the average time spent on important components, including voxel allocation, ray-voxel intersection test and volume rendering, etc. The experiment is performed on a single NVidia RTX 3090 video card. The results are listed in~\autoref{tab:time}. Please note that tracking and mapping are profiled on a per-iteration basis. As can be seen, our voxel manipulation functions have no significant impact on the running time of the reconstruction pipeline. Depending on the scene complexity, our method can take around $150$-$200$ ms to track a new frame and $450$-$550$ ms for the joint frame and map optimization. In a typical setting, our system can run $5hz$ for tracking, and $2hz$ for optimization. However, for more challenging scenarios, the system might run slower. 

As explained before, our sparse voxel structure allows us to only allocate voxels occupied by objects and surfaces, which is often only a fraction of the entire environment. We profile our system as well as NICE-SLAM for memory consumption of implicit decoders and voxel embeddings on the Replica office-0 scene. the results are listed in~\autoref{tab:param}. Please note that NICE-SLAM uses $4$ layers of densely populated voxel grids while we only use one. It can be clearly seen that our method can achieve better reconstruction accuracy while using significantly less memory.

\begin{table}[t]
\caption{Memory consumption for implicit features.}
\label{tab:param}
  \scriptsize%
	\centering%
  \begin{tabu}{%
	l%
	*{2}{c}%
	*{2}{c}%
	}
  \toprule
  Method & Decoder & Embedding \\
  \midrule
    Ours & 1.04MB & 0.149MB \\
    NICE-SLAM & 0.22MB & 238.88MB \\
  \bottomrule
  \end{tabu}%
\vspace{-1em}
\end{table}

\section{Applications}
\label{sec:app}

Our Vox-Fusion can not only estimate accurate camera poses in cluttered scenes, but also obtain dense depth maps with fine geometric details and render realistic images through our differentiable rendering. Therefore, it can be applied to many AR and VR applications. For AR applications, our method allows us to place arbitrary virtual objects into reconstructed scenes, and accurately represent the occlusion relationship between real and virtual contents using the rendered depth maps. We show examples of rendered images and AR demo images in~\autoref{fig:ar_images}. 
As can be seen, our dense scene representation allows us to handle occlusion between different objects very well in the AR demo. 

For VR applications, apart from providing accurate camera tracking result, the realistic rendering ability can be used in free-view virtual scene traveling. Our voxel-based method also makes scene editing much easier since we can remove part of the scene by simply deleting its supporting voxels and the associated feature embeddings. The explicit voxel representation can also be used to perform fast collision detection. We can also control the level of details by splitting and refining feature embeddings as introduced in~\cite{liu:2020:nsvf}. 

\section{Conclusion}
\label{sec:limit}

We propose Vox-Fusion, a novel dense tracking and mapping system built on voxel-based implicit surface representation. Our Vox-Fusion system supports dynamic voxel creation, which is more suitable for practical scenes. Moreover, we design a multi-process architecture and corresponding strategies for better performance. Experiments show that our method achieves higher accuracy while using smaller memory and faster speed. Currently, our Vox-Fusion method cannot robustly handle dynamic objects and drift in long-time tracking. We consider these as potential future works. 

\acknowledgments{This work was partially supported by the National Key Research and Development Program of China under Grant 2020AAA0105900.}


\bibliographystyle{abbrv-doi}

\bibliography{main}
\end{document}